\documentclass{article} % For LaTeX2e
\usepackage{iclr2024_conference,times}

\usepackage{amsmath,amsfonts,bm}

\def\eqref#1{equation~\ref{#1}}
\def\1{\bm{1}}

\DeclareMathAlphabet{\mathsfit}{\encodingdefault}{\sfdefault}{m}{sl}
\SetMathAlphabet{\mathsfit}{bold}{\encodingdefault}{\sfdefault}{bx}{n}

\usepackage{natbib}
\usepackage[hidelinks]{hyperref}
\usepackage{url}
\usepackage{booktabs}
\usepackage{subfigure}
\usepackage{multirow}
\usepackage{colortbl}
\usepackage{makecell}
\usepackage{graphicx}
\usepackage{wrapfig}

\title{Towards Green AI in Fine-tuning Large Language Models via Adaptive Backpropagation}

\author{Kai Huang\footnotemark[2] , \ Hanyun Yin\footnotemark[4] , \ Heng Huang\footnotemark[3] \ \ \& Wei Gao\footnotemark[2] \\
University of Pittsburgh\footnotemark[2] , University of Maryland, College Park\footnotemark[3] \\
University of Science and Technology of China\footnotemark[4] \\
\texttt{k.huang@pitt.edu}, \ \ \texttt{ykissgoodbye@gmail.com}, \ \ \texttt{heng@umd.edu}, \ \ \texttt{weigao@pitt.edu}
}

\iclrfinalcopy % Uncomment for camera-ready version, but NOT for submission.
\begin{document}

\maketitle

	\vspace{-0.1in}
\begin{abstract}
	\vspace{-0.1in}
    Fine-tuning is essential to adapting pre-trained large language models to downstream applications. With the increasing popularity of LLM-enabled applications, fine-tuning has been performed intensively worldwide, incurring a tremendous amount of computing costs that correspond to big carbon footprint and environmental impact. Mitigating such environmental impact directly correlates to reducing the fine-tuning FLOPs. Existing fine-tuning schemes focus on either saving memory or reducing the overhead of computing weight updates, but cannot achieve sufficient FLOPs reduction due to their ignorance of the training cost in backpropagation. To address this limitation, in this paper we present \emph{GreenTrainer}, a new technique that minimizes the FLOPs of LLM fine-tuning via adaptive backpropagation, which adaptively selects the most appropriate set of LLM tensors for fine-tuning based on their importance and backpropagation cost in training. Experiment results show that GreenTrainer can save up to 64\% training FLOPs compared to full fine-tuning, without any noticeable accuracy loss. Compared to the existing schemes such as Prefix Tuning and LoRA, GreenTrainer can achieve up to 4\% improvement of model accuracy, with on-par FLOPs reduction. 
\end{abstract}

\vspace{-0.2in}
\section{Introduction}
\vspace{-0.1in}

Large language models (LLMs) are used as foundational tools in generative AI. To be used in downstream applications, a pre-trained LLM needs to be fine-tuned using the specific application data \citep{devlin2018bert}. Intuitively, fine-tuning is less computationally expensive than pre-training due to the smaller amount of training data, but it may result in significantly high energy consumption and carbon footprint when being intensively performed worldwide. Enabled by the democratization of open-sourced LLMs \citep{candel2023h2ogpt} and convenient APIs of operating these LLMs \citep{ott2019fairseq, wolf2019huggingface}, even non-expert individuals can easily fine-tune LLMs for model performance enhancement or personalization \citep{scialom2022fine,wang2023tackling}. For example, when a LLaMA-13B model \citep{touvron2023llama} is fine-tuned by 10k users using A100-80GB GPUs, such fine-tuning consumes 6.9$\times$ more GPU hours than pre-training a GPT-3 model \citep{brown2020language} with 175B parameters. The amount of energy consumed by such fine-tuning is comparable to that consumed by some underdeveloped countries, and the amount of carbon footprint is equivalent to 1000$\times$ of that produced by a New York-San Francisco flight \citep{aiindex2023}.

%\vspace{-0.05in}
Mitigating such environmental impact towards Green AI directly correlates to reducing the number of floating operations (FLOPs) of fine-tuning, which represents the amount of computational operations and hence energy consumption in training \citep{schwartz2020green,huang2023elastictrainer}. Most existing techniques of optimizing LLM fine-tuning, however, are limited to reducing the memory consumption rather than FLOPs \citep{malladi2023fine, liao2023make}. Some other methods reduce FLOPs by only fine-tuning certain types of model parameters such as bias \citep{zaken2021bitfit}, LayerNorm and output layer weights \citep{lu2021pretrained}, but they impair the model's expressivity and are only applicable to simple non-generative learning tasks. Instead, researchers suggested keeping the original model parameters frozen but injecting additional trainable parameters to the input \citep{lester2021power, liu2022p} or internal layers \citep{li2021prefix, hu2023llm,huang2023modality}. Recent LoRA-based methods \citep{hu2021lora, zhang2023adaptive} further reduce the overhead of computing weight updates for these injected parameters via low-rank approximation. These methods can minimize the model's accuracy loss on generative tasks. However, they still need to compute the activation gradients through the whole model and their FLOPs reduction is hence limited, because the computations of weight updates are only 25\%-33\% of the total training FLOPs.

%the backpropagation cost should be modeled as a constraint that is incorporated into trainable portion selection. 
Besides computing weight updates, FLOPs in training are also produced in i) forward propagation and ii) backward propagation of activation gradients. Since complete forward propagation is essential to calculate the training loss, we envision that the key to further FLOPs reduction is to take the backpropagation cost of activation gradients, which is $>$33\% of the total training FLOPs, into account and selectively involve only the most appropriate model structures in backpropagation. The major challenge, however, is that selective training will possibly bring model accuracy loss. We minimize the accuracy loss is by adapting such selection in backpropagation to a flexible objective of FLOPs reduction, determined by the carbon footprint in energy supply. For example, when such carbon footprint is low due to insertion of renewable energy, using a lower objective of FLOPs reduction can involve more model structures in training and retain the training accuracy. High carbon footprint, instead, leads to a higher objective of FLOPs reduction for better embracing Green AI.

\begin{wrapfigure}{R}{0.4\textwidth}
	\centering
	\vspace{-0.2in}
	\includegraphics[width=0.4\textwidth]{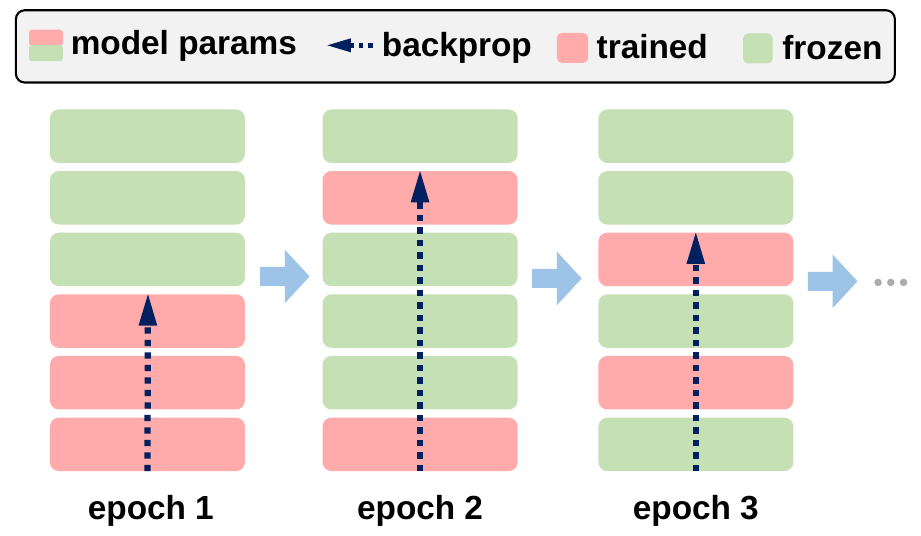}
	\vspace{-0.25in}
	\caption{\label{fig:adaptive_bp} GreenTrainer adaptively selects the most appropriate portion of LLM model for fine-tuning}
	\vspace{-0.1in}
\end{wrapfigure}

In this paper, we present \emph{GreenTrainer}, a new technique that realizes adaptive backpropagation for efficient LLM fine-tuning with the minimum accuracy loss. As shown in Figure \ref{fig:adaptive_bp}, given an objective of FLOPs reduction, GreenTrainer adaptively selects the set of trainable neural network (NN) tensors in each epoch, based on evaluation of tensors' importance in training. Such importance evaluation is difficult because NN tensors do not directly associate with input data variables or intermediate features, and most attribution techniques \citep{sundararajan2017axiomatic, hesse2021fast} that evaluate feature importance are not applicable. Popular importance metrics, including SNIP \citep{lee2018snip} and Fisher \citep{liu2021group}, are mainly used in NN pruning to quantify the importance of model weights at their current values, but they cannot quantify the importance of weight updates on a tensor to reducing the training loss.  Classic metrics based on exact accuracy contribution \citep{lin2022device}, weight updates' magnitudes \citep{li2016pruning}, or random perturbations \citep{breiman2001random}, on the other hand, are either inaccurate or computationally expensive for LLMs. Instead, our approach adopts a similar rationale with the existing attribution and pruning metrics, and quantifies the contribution of each tensor update to the training loss via first-order Taylor expansion over the training loss. In this way, we ensure that the selected tensors can make the maximum contribution to reducing the training loss.

Another challenge is how to precisely profile the training FLOPs. Due to interdependency between tensors, their total FLOPs in training is not equal to the summation of their individual FLOPs. Such interdependency is determined by the backpropagation characteristics of  NN operators in each tensor, but existing FLOPs models cannot link NN operators to tensors based on the computing flow of backpropagation. Some existing work \citep{kwon2023tinytrain} only incorporates the layer-wise forward FLOPs into tensor selection, but ignores the computation dependency between layers in backpropagation. To tackle this challenge, we rigorously model the cross-tensor dependencies in profiling their backpropagation FLOPs. Based on this model, we develop a dynamic programming (DP) algorithm to find the nearly optimal tensor selection from an exponential number of possibilities (e.g., $2^{515}$ for 515 tensors in OPT-2.7B model \citep{zhang2022opt}). Therefore, GreenTrainer can make sure that the given objective of FLOPs reduction can be met in most cases.

% To tackle this challenge, we build a new FLOPs model that incorporates the relations between tensors and NN operations into FLOPs profiling. Based on this model, we develop a dynamic programming (DP) algorithm to find the nearly optimal tensor selection from an exponential number of possibilities (e.g., $2^{515}$ for 515 tensors in OPT-2.7B model \citep{zhang2022opt}).

%, via adaptive control of backpropagation cost in training
%To our best knowledge, GreenTrainer is the first work that adapts the training FLOPs reduction in LLM fine-tuning to the need of Green AI. Our detailed contributions are as follows:
%\begin{itemize}
%\item \textbf{Tensor FLOPs Profiling (Section 3.1).} We build new FLOPs models that allow precise profiling of tensor selection’s training FLOPs for transformer-based LLMs.
%\item \textbf{Efficient Tensor Importance Evaluation (Section 3.2).} We developed a runtime evaluation process of tensor importance with low memory cost, high accuracy, and high adaptability.
%\item \textbf{Efficient DP for Tensor Selection (Section 3.3).} Our lightweight DP algorithm can decide the optimal selection of tensors at runtime that maximizes the training loss reduction.
%\end{itemize}

We evaluated GreenTrainer with three open-sourced LLMs, namely OPT \citep{zhang2022opt}, BLOOMZ \citep{muennighoff2022crosslingual} and FLAN-T5 \citep{chung2022scaling}, on text generation datasets including SciTLDR \citep{cachola2020tldr} and DialogSum \citep{chen2021dialogsum}. Our  results show that GreenTrainer can save up to 64\% training FLOPs compared to full LLM fine-tuning, without any noticeable accuracy loss. In some cases, GreenTrainer can even improve the model accuracy compared to full fine-tuning, by removing model redundancy and overfitting. Compared to existing techniques such as Prefix Tuning \citep{li2021prefix} and LoRA \citep{hu2021lora}, GreenTrainer improves the model accuracy by 4\% with the same amount of FLOPs reduction, and also provides users with the flexibility to balance between the training accuracy and cost depending on the needs of Green AI.
%We don't choose non-generative tasks (e.g., extractive QA and sentiment analysis) since they are rarely applied to existing LLMs in practice.

% lora converts the original weights into a more computationally efficient version via low-rank decomposition
% but according to Green AI proposal, reducing the number of parameters cannot directly reduce CO2 emission. Instead, training FLOPs is a more effective objective for minimization.

\vspace{-0.05in}
\section{Background \& Motivation}

\vspace{-0.05in}
\subsection{Transformer Architectures for Text Generation}
\vspace{-0.05in}

Current LLMs are stacked by transformer blocks \citep{vaswani2017attention}, each containing a Multi-Head Attention (MHA) layer, LayerNorms \citep{ba2016layer}, and a Feed-Forward Network (FFN). Given an input sequence $X \in \mathbb{R}^{n\times d}$ with $n$ tokens, the MHA projects tokens into a $(Q, K, V)$ space $h$ times, using $h$ suites of trainable projectors $(W_Q^{(i)}, W_K^{(i)}, W_V^{(i)})_{i = 1, ..., h}$. Each projection $f_i: \mathbb{R}^{n\times d} \to \mathbb{R}^{n\times \frac{d}{h}}$ is defined as $Q_i, K_i, V_i = XW_Q^{(i)}, XW_K^{(i)}, XW_V^{(i)}$. The output $(Q_i, K_i, V_i)$ then performs attention mechanisms to produce $O_i$ by weighting $V_i$ with the attention scores between $Q_i$ and $K_i$. The MHA's final output is obtained by concatenating each $O_i$, following a linear projection $g: \mathbb{R}^{n\times d} \to \mathbb{R}^{n\times d}$ with a trainable projector $W_o$:
\vspace{-0.05in}
\begin{align}\label{eq:attention}
        O_i &= \mathrm{Softmax}\left(Q_i K_i^{\top} / \sqrt{d/h}\right)V_i, & \mathrm{MHA_{out}} = \mathrm{Concat}(O_1, O_2, ..., O_h) W_o.
        \vspace{-0.05in}
\end{align}
%Due to their auto-regressive nature, LLMs can only generate a single output token in each forward pass, which is inefficient in training.
To improve the training efficiency, LLMs adopt the teacher-forcing method \citep{lamb2016professor} to generate the entire sequence of output tokens in a single forward pass. Specifically, causal masks are applied to MHA's attention scores, so that each output token can be predicted from the label tokens at previous positions. With this technique, when being fine-tuned, LLMs can be trained in a standard way like any feed-forward models.

\vspace{-0.05in}
\subsection{The Need for Adaptive Backpropagation}
\vspace{-0.05in}
% dataset scale, model variants
% pre-trained LLMs have strong commonsense knowledge and zero-shot adaptability (fine-tuning is still better)
% we have seen fine-tuning only a small portion of LLM can achieve good accuracy.
% identify important portions and control FLOPs reduction
%By stacking a sufficient number of large transformer blocks, pre-trained LLMs can capture general language patterns and world knowledge. However, w
When being fine-tuned for a downstream task, LLMs are usually over-parameterized, because only part of the world knowledge that they learned from pre-training is useful for the target task. In these cases, only involving some of the model's substructures into fine-tuning could have little impact on the model accuracy, but significantly reduces the amount of computations. 

\begin{table}[ht]
	\vspace{-0.1in}
	\centering
	{\fontsize{7}{9}\selectfont
		\vspace{0.1in}
		\begin{tabular}{lrrrr}
			\toprule
			\multirow{2}{*}{\makecell{\textbf{Trainable} \\ \textbf{substructure}}} & \multicolumn{2}{c}{\textbf{OPT-2.7B}} & \multicolumn{2}{c}{\textbf{FLAN-T5-3B}} \\
			\cmidrule(lr){2-3} \cmidrule(lr){4-5}
			& \textbf{FLOPs ($\times 10^{15}$)}  & \textbf{Acc. (\%)} & \textbf{FLOPs ($\times 10^{15}$)}  & \textbf{Acc. (\%)} \\
			All params        & 262.0  & 23.6   & 135.7 & 46.5   \\
			Last 2 layers        & 181.6 (31\%$\downarrow$) & 20.8   & 46.1 (66\%$\downarrow$) & 39.2   \\
			Decoder prefix      & 174.7 (33\%$\downarrow$) & 13.4   & 55.3 (60\%$\downarrow$) & 37.6   \\
			$(W_Q, W_V)$  & 174.7 (33\%$\downarrow$) & 23.8   & 90.5 (33\%$\downarrow$) & 44.7   \\
			\bottomrule
	\end{tabular}}
	\caption{Fine-tuning different substructures of OPT-2.7B and FLAN-T5-3B LLMs on the DialogSum dataset (ROUGE-1 score on the test set is used as the accuracy metric)}
	\label{tab:opportunities}
        %\vspace{-0.1in}
\end{table}

Existing work has made attempts with fixed selections of some NN components, such as the last 2 layers, decoder prefixes \citep{li2021prefix}, and linear projectors $(W_Q, W_V)$ \citep{hu2021lora}, in fine-tuning. However, due to the interdependencies of NN parameters \citep{jin2020does}, such fixed selections will significantly impair the model accuracy. As shown in Table \ref{tab:opportunities}, solely fine-tuning either the last 2 layers or decoder prefixes leads to up to 10\% accuracy drop. The reason is that nearby NN substructures with interdependencies on the fixed selections are excluded from fine-tuning, and hence become inconsistent with those selected substructures. Increasing the density of selection, such as including all the linear projectors $(W_Q, W_V)$, could mitigate the model accuracy loss, but can save at most 33\% FLOPs due to backpropagating activation gradients through transformer blocks. Naive methods of dynamic selections, such as expanding the trainable portion from the last layer, have the similar limitation.

The deficiency of these existing methods motivates us to enforce more flexible and adaptive selection of LLM substructures in backpropagation. In GreenTrainer, we develop a tensor importance metric that incorporates parameter dependencies to evaluate how fine-tuning each tensor contributes to the trained model's accuracy at runtime. Knowledge about such tensor importance, then, allows us to achieve the desired FLOPs reduction while maximizing the model accuracy.
 
% Based on this intuition, existing work provides several options for fine-tuning LLM substructures. However, their selections are not optimized for both accuracy and FLOPs reduction.

%[Not sure if we should mention the need for "runtime" selection here.]

\begin{figure}[ht]
	\centering
	\vspace{-0.1in}
	\includegraphics[width=0.7\linewidth]{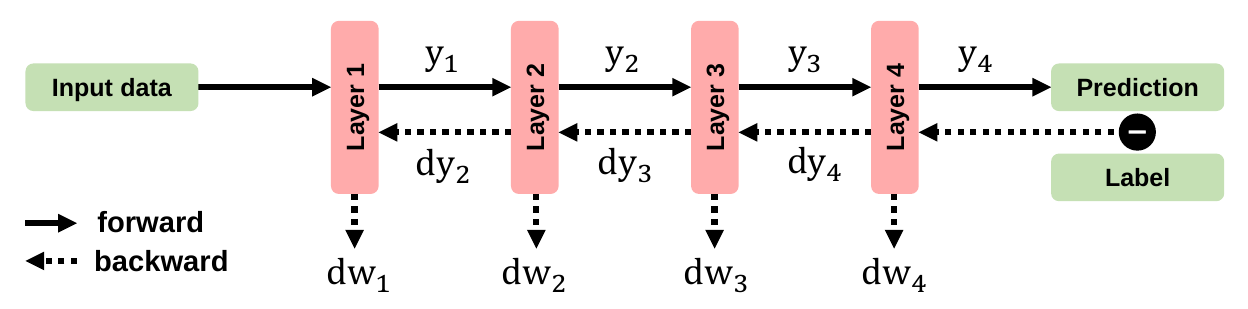}
	\vspace{-0.1in}
	\caption{Backpropagation of a 4-layer dense NN}
	%	\vspace{-0.05in}
	\label{fig:bp_time_model}
	\vspace{-0.05in}
\end{figure}

\vspace{-0.05in}
\subsection{FLOPs Model of Backpropagation}
\vspace{-0.05in}
% tightly connect to the previous paragraph
% Describe how backpropagation is generally done in neural networks and how training FLOPs can be calculated for natural language generation training.
The design of GreenTrainer relies on proper calculation of the selected model substructures' backpropagation FLOPs, which can be decomposed into two parts using the chain rule. For example, as shown in Figure \ref{fig:bp_time_model}, when training a 4-layer dense NN without bias, each layer computes \emph{i)} $\mathrm{dy_i}$ as the loss $L$'s gradient w.r.t the activation $y_i$, and \emph{ii)} $\mathrm{dw_i}$ as the loss gradient w.r.t weight $W_i$, such that
\begin{align}\label{eq:dense_bp}
        \mathrm{dy_i} &= \frac{\partial L}{\partial y_i} = \frac{\partial L}{\partial y_{i+1}} W_i^{\top} = \mathrm{dy_{i+1}} W_i^{\top}, & \mathrm{dw_i} = \frac{\partial L}{\partial W_i} = y_i^{\top} \frac{\partial L}{\partial y_{i+1}} = y_i^{\top} \mathrm{dy_{i+1}},
\end{align}
and the corresponding amounts of FLOPs for computing $\mathrm{dy_i}$ and $\mathrm{dw_i}$ are $t_{dy_i}$ and $t_{dw_i}$, respectively.

$(\mathrm{dy_i}, \mathrm{dw_i})$ can be computed from $(\mathrm{dy_{i+1}}, \mathrm{dw_{i+1}})$. In particular, even if a layer is not selected in fine-tuning, it still needs to compute and pass error gradients ($\mathrm{dy_i}$) to the downstream layers. Hence, the amount of computations in backpropagation does not only depend on the selected layers, but also depends on some unselected layers. For example, if only Layer 2 is trainable, the total FLOPs for backpropagation will be decided by the cost of computing $\mathrm{dw_{2}}$, $\mathrm{dy_{3}}$ and $\mathrm{dy_{4}}$. Due to the generality of the chain rule, such rationale of FLOPs calculation is also applicable to other types of NN layers. 
%Both $\mathrm{dy_i}$ and $\mathrm{dw_i}$ take considerable computation and either cannot be omitted. 

Based on this rationale, we can construct FLOPs models for LLM substructures. The layer-level model is coarse-grained and can lead to inaccurate tensor selection. Some important parameters may be unselected due to other unimportant ones in the same layer. In GreenTrainer, we use tensor-level granularity for such selection, which can be well-supported by tensorized NN libraries (e.g., TensorFlow \citep{abadi2016tensorflow} and PyTorch \citep{paszke2019pytorch}). Weight-level selection, although more fine-grained, is too computationally expensive due to the requirement of fine-grained indexing.

\vspace{-0.1in}
\section{Method}
\vspace{-0.1in}

%\noindent\textbf{Problem formulation.}
To reduce the FLOPs of LLM fine-tuning, an intuitive problem formulation is to minimize the FLOPs while achieving the desired model accuracy. However, it is hard to determine a proper accuracy objective in advance, because some accuracy objectives may require very intensive training and the accuracy that we can achieve with our FLOPs budget cannot be pre-estimated before training. Instead, we maximize the training loss reduction while achieving the desired FLOPs reduction:
%\vspace{-0.05in}
\begin{align}\label{eq:general_formulation}
    \max{\Delta_{loss}(\bm{m})} \ \ \ \text{s.t.} \ \ T_{selective}(\bm{m}) \le \rho T_{full},
%    \vspace{-0.05in}
\end{align}
where $\bm{m}$ is a binary vector to be solved for tensor selection. $\bm{m}$ parameterizes both the loss reduction ($\Delta_{loss}$) and per-batch FLOPs of training ($T_{selective}$), and $T_{selective}$ is constrained within a user-specified ratio ($\rho$) of the FLOPs of fine-tuning the whole model ($T_{full}$). For example, $\rho=0.5$ means that the FLOPs of fine-tuning should be at most 50\% of that in fine-tuning the whole model. In practice, the value of $\rho$ can either be preset or adjusted at runtime in any stage of training.

To identify each tensor's contribution in fine-tuning, we model $\Delta_{loss}(\bm{m})$ as the aggregated importance of selected tensors, and calculate the FLOPs incurred by selected tensors using the FLOPs model of backpropagation in Section 2.3. With this model, Eq. (\ref{eq:general_formulation}) can be rewritten as:
%\vspace{-0.05in}
\begin{align}\label{eq:detailed_formulation}
    \max{\ \Delta_{loss}(\bm{m})} \ \ \ \ \ \text{s.t.} \ \ T_{fp} + \bm{m} \cdot \bm{t}_{dw} + \sigma(\bm{m}) \cdot \bm{t}_{dy} \le \rho T_{full},
%    \vspace{-0.05in}
\end{align}
where $T_{fp}$ indicates the per-batch FLOPs of the forward pass, and each pair of variables in $(\bm{t}_{dy}, \bm{t}_{dw})$ represents the FLOPs of computing $(\mathrm{dy}, \mathrm{dw})$ for the corresponding tensor, respectively. Given a binary selector $\bm{m}$, $\sigma(\bm{m})$ incorporates all the tensors along the backward pass that contribute to the FLOPs of fine-tuning, by involving in passing the error gradients ($\mathrm{dy}$). For example, if $\bm{m}=[0,0,1,0,1,0,0]$, all the tensors that are in deeper layers than the selected tensors are involved in passing the error gradients, and hence $\sigma(\bm{m})=[0,0,1,1,1,1,1]$.

To ground this formulation and solve $\bm{m}$, GreenTrainer consists of three key components: \emph{(i) Tensor FLOPs Profiling}, which calculates the FLOPs of all NN tensors (i.e., $\bm{t}_{dy}$ and $\bm{t}_{dw}$) prior to training; \emph{(ii) Tensor Importance Evaluation}, which quantifies the contribution of updating each NN tensor to the training quality at runtime; \emph{(iii) Tensor Selector}, which grounds the tensor selection problem using tensors' FLOPs and importances, and provides solutions via dynamic programming at runtime.

\vspace{-0.05in}
\subsection{Tensor FLOPs Profiling}
\vspace{-0.05in}
Standard NN profilers, such as Torch Profiler \citep{paszke2019pytorch}, can measure the execution FLOPs of individual NN operators such as matrix multiplication and convolution. However, it cannot be directly linked to NN tensors that participate in these operations. When a set of tensors is trained, the training FLOPs of backpropagation are not equal to the summation of individual tensors' FLOPs.

\begin{figure}[ht]
	\centering
	\vspace{-0.15in}
	\includegraphics[width=0.85\linewidth]{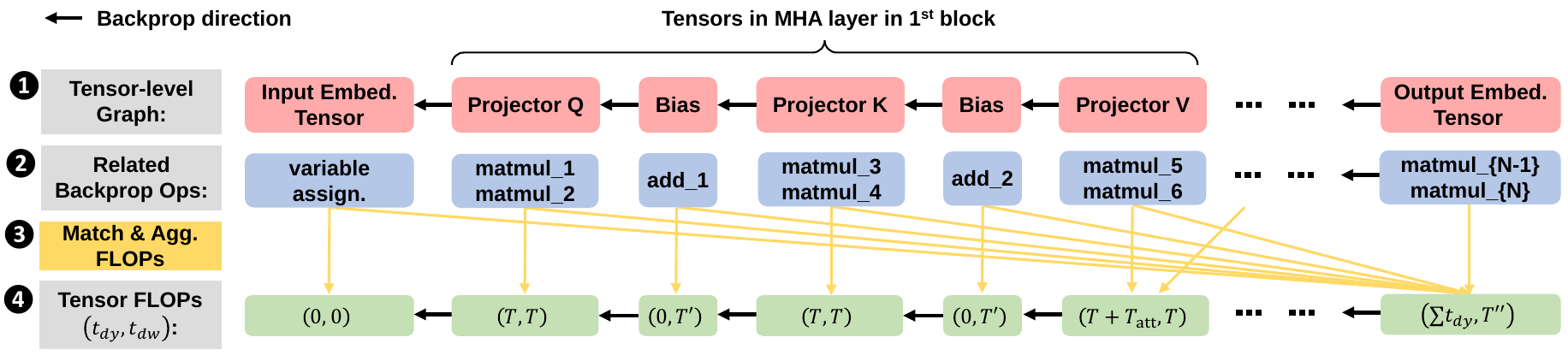}
	\vspace{-0.1in}
	\caption{An sample workflow of tensor FLOPs profiling}
	\vspace{-0.05in}
	\label{fig:flops_profiling}
	\vspace{-0.1in}
\end{figure}

To address this limitation, our approach consists of two steps. First, we convert the layer-based NN structure of LLMs into a tensor-level computing graph, which retains the execution order of all tensors' involvements in training. Then, we extract the related backpropagation operators of each tensor, and derive each tensor $i$'s FLOPs in backpropagation ($t_{dy_i}$ and $t_{dw_i}$) by matching and aggregating the FLOPs of these NN operators. For example in Figure \ref{fig:flops_profiling}, the training of each linear projector ($Q$, $K$ and $V$) in an MHA layer should be executed after its corresponding bias tensor's training. Training each linear projector, then, will involve two matrix multiplication operators, whose FLOPs in backpropagation will be aggregated. We categorize such rules of matching and aggregation by the type of LLM layers where tensors are located, as described below. A specific example about such tensor FLOPs profiling on the OPT-2.7B model is provided in Appendix \ref{appendix:profiling}.

%Describe how the time model is constructed in detail regarding different transformer's building blocks, e.g., Multi-Head Attention, LayerNorm, FFN, Output Embedding Layer.

\noindent\textbf{Input \& output embedding layers.}
The input embedding layer contains a trainable embedding tensor that maps each raw token into a dense representation. Given the activation gradient $\rm{dy_{i+1}}$ from upstream layers, deriving the update of this tensor only involves variable assignment, and we can safely consider $t_{dw_i}\approx 0$ for any tensor $i$. If a raw token is mapped to the $k$-th vector in the embedding tensor during the forward pass, then during backpropagation, $\rm{dy_{i+1}}$ from the upstream will be only assigned to $k$-th row of $\rm{dw_{i}}$, such that $\rm{dw_{i}}[s] = \rm{dy_{i+1}}$ if $s=k$, otherwise $\rm{dw_{i}}[s]=0$. Since the input layer doesn't propagate activation gradients, we can also conclude that its $t_{dy}$ is 0.

%by multiplying its trainable tensor with the token vector
Reversely, the output embedding layer projects each token back to the probability space. Intuitively, its $(t_{dy}, t_{dw})$ can be derived in the same way as we did for the dense layer in Eq. (\ref{eq:dense_bp}). However, in most LLMs, the output embedding layer shares the same trainable tensor with the input embedding layer. This implies that if the output embedding is trainable, then the input embedding will also be involved in training. Hence, all the $t_{dy}$ from LLM's output, up to the input embedding layer, should be accumulated to $t_{dy}$ of the output embedding tensor, while its $t_{dw}$ remains unchanged.

\noindent\textbf{Multi-Head Attention (MHA) layer.}
An MHA layer contains multiple linear projectors as trainable tensors, and their FLOPs in training can be derived in the same way as we did with the dense layer in Eq. (\ref{eq:dense_bp}). Some LLMs (e.g., OPT) also include bias as another type of trainable tensor after such projection. In this case, based on the chain rule, the backpropagation of bias is computed as $\rm{dy_i}=\rm{dy_{i+1}}$ and $\rm{dw_i} = \bm{1}^{\top} \rm{dy_{i+1}}$, indicating that $t_{dy}$ for bias is 0 since ${\rm{dy_i}}$ is identically passed from $\rm{dy_{i+1}}$. $t_{dw}$ of bias can be derived as the FLOPs of adding up elements in ${\rm{dy_{i+1}}}$ along every feature channel. The attention mechanism in Eq. (\ref{eq:attention}) is backpropagated prior to the projectors. If any of these projectors are involved in training, the attention's backpropagation FLOPs must be also calculated, and we accumulate such FLOPs to the corresponding projector tensor ($W_V$)'s $t_{dy}$.

\noindent\textbf{LayerNorm.}
Given a token, LayerNorm first normalizes its features and uses two trainable tensors $\gamma$ and $\beta$ to element-wise multiply with and add to the token, respectively. The operations of multiplication and addition are similar to those in the dense layer, and so its FLOPs can be calculated in the similar way. However, the backpropagation FLOPs of normalization operators should be accumulated to the previous tensor's $t_{dy}$. If any tensors in the previous layers are trained, the FLOPs of propagating the normalization operators should be also included in the FLOPs of the current layer.

\noindent\textbf{Feed-Forward Network (FFN).}
In the FFN, there is a nonlinear activation function between two dense layers. Following the same method of calculating LayerNorm's FLOPs, we accumulate the FLOPs of propagating through this activation function to the bias tensor's $t_{dy}$ in the first dense layer.

\vspace{-0.1in}
\subsection{Tensor Importance Evaluation}
\vspace{-0.05in}
%\noindent\textbf{Gradient-based importance metric.}
A tensor's importance in training can be estimated as the summation of the importances of all its weights. In training, since the model weights are iteratively updated to minimize the training loss, an intuitive approach to evaluating the importance of a weight update in a given iteration is to undo this update and check how the training loss increases back as $\Delta L = L(w) - L(w + \Delta w)$, so that a higher value of $\Delta L$ means this update is more important and the weight should be selected. However, computing $\Delta L$ for every weight is expensive. Instead, we estimate the importance of all weights in one shot by smoothing the undo operation described above and computing the loss gradients with respect to the updates that correspond to all the weights. Letting the multiplicative $\bm{c} \in [0,1]^M$ denote the undo operation for all the $M$ weights, we can compute the loss gradient as 
%\vspace{-0.05in}
\begin{align}\label{eq:soft_undo}
    -\frac{\partial L(\bm{w} + \bm{c} \odot \Delta \bm{w})}{\partial \bm{c}} = -\left. \Delta \bm{w} \odot \frac{\partial L(\bm{u})}{\partial \bm{u}} \right\vert_{\bm{u}=\bm{w} + \bm{c} \odot \Delta \bm{w}},
%\vspace{-0.05in}
\end{align}
where $\odot$ denotes element-wise multiplication. When $\bm{c} = \bm{0}$, Eq. (\ref{eq:soft_undo}) becomes an importance vector over all weights. Since the loss gradient is parameterized by all weights, the weight importances calculated in this way implicitly incorporate the impact of weight dependencies. A tensor $k$'s importance is then calculated as
\vspace{-0.05in}
\begin{align}\label{eq:importance}
I_k &= -\sum\nolimits_{i} \Delta w_i^{(k)} \partial L/\partial w_i^{(k)}.
%& \widehat{I}_k = I_k / \max(|I_1|, |I_2|, ...).
\vspace{-0.05in}
\end{align}
In some cases, when the training process encounters divergence, the values of gradients and calculated tensor importances in Eq. (\ref{eq:importance}) could be very large, eventually leading to overflow when using these importance values for deciding tensor selection in Eq. (\ref{eq:detailed_formulation}). To address this issue, we could further scale all the tensor importance by the maximum amplitude to improve numerical stability.

%The importance of a tensor $k$, then, is a summation of all its weights' importance.

%Intuitively, if the training adopts naive gradient descent algorithms, each weight update $\Delta w_i$ should be equivalent to $-\alpha {\partial L}/{\partial w_i}$ where $\alpha$ is the learning rate, and Eq. (\ref{eq:importance}) can hence be simplified [how? provide some details!]. However, modern NN training usually includes schedulers (e.g., cosine decay) and advanced optimizers with momentum and runtime scaling, such as Adam \citenum{kingma2014adam} and AdamW \citenum{loshchilov2017decoupled}. In this case, a more general expression of the weight update is
%\begin{align}\label{eq:delta_w}
%    \Delta w_i^{(k)} = -{\rm{\mathbf{Scheduler}}}(\alpha) \cdot {\rm{\mathbf{Optimizer}}}\left({\partial L}/{\partial w_i^{(k)}}\right).
%\end{align}

%In most existing NN libraries such as PyTorch and TensorFlow, it is not allowed to retrieve the runtime status of such scheduler and optimizer about the calculation of Eq. (\ref{eq:delta_w}). Instead, we run a probing iteration to obtain the weight update by $\Delta \bm{w}=\bm{w}_{t+1} - \bm{w}_t$. On the other hand, to prevent this probing iteration from affecting the training dynamics, we cache the original weights $\bm{w}_t$ before probing, and restore them afterward.

\begin{figure}[ht]
	\centering
	\vspace{-0.15in}
	\hspace{-0.25in}
	\subfigure[Subproblem definition] { 
		\includegraphics[width=0.38\textwidth]{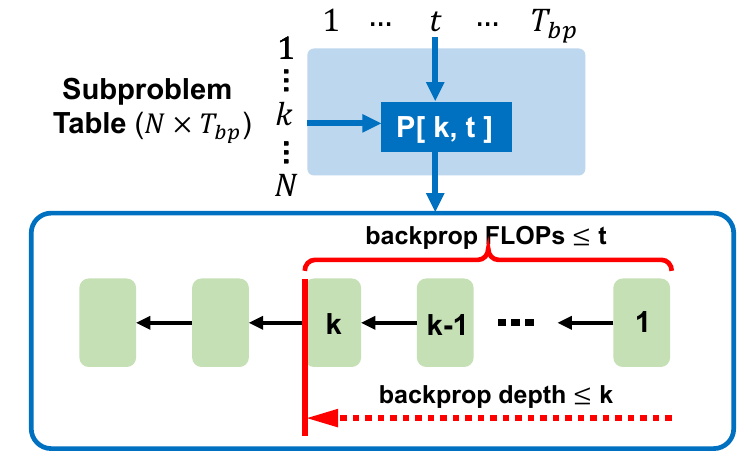}
		\label{fig:subproblem_def}
	}
	\hspace{0.1in}
	\subfigure[Finding recurrence relations] { 
		\includegraphics[width=0.38\textwidth]{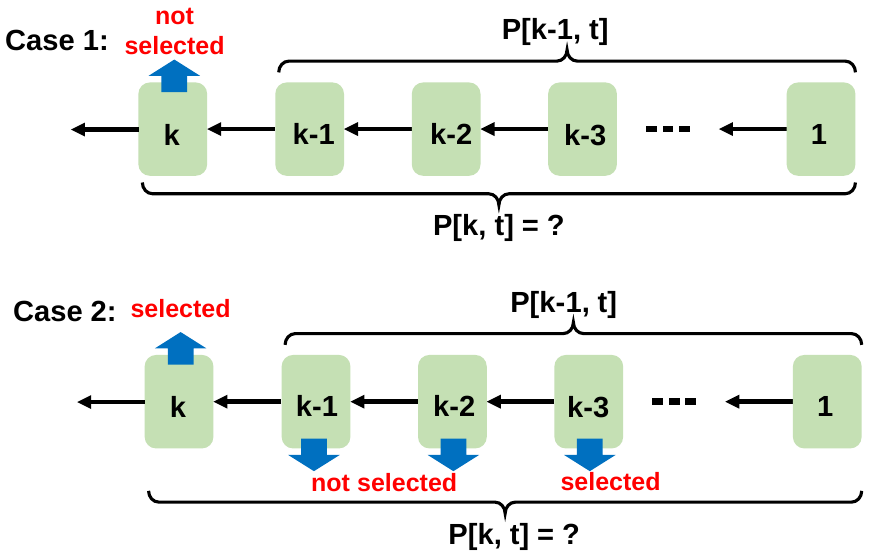}
		\label{fig:recurrence_relation}
	}
	\hspace{-0.4in}
	\vspace{-0.15in}
	\caption{Solving the tensor selection problem using DP}
	\label{fig:dp}
	\vspace{-0.15in}
\end{figure}

\vspace{-0.05in}
\subsection{Tensor Selection}
\vspace{-0.05in}
%transform $\rho$ to the reduction ratio of backpropagation cost.
%Put the same content as Dynamic Programming and Computation Reduction in ElasticTrainer, except replacing pure text description with pseudo code.
Since Eq. (\ref{eq:detailed_formulation}) is a nonlinear integer programming problem and hence NP-hard, in GreenTrainer we seek for an approximate solution using dynamic programming (DP). We decompose the whole problem into subproblems constrained by different depths of backpropagation. These subproblems can be sequentially solved from the one with the smallest depth, by using their recurrence relations. 

%Since we only focus on backpropagation FLOPs, we can equivalently convert the original objective $\rho$ w.r.t the (forward + backpropagation) FLOPs into the one only w.r.t the backpropagation FLOPs as $\rho_{bp}$. The equivalent constraint and $\rho_{bp}$ are: 
%\begin{align}
 %   & \bm{m} \cdot \bm{t}_{dw} + \sigma(\bm{m}) \cdot \bm{t}_{dy} \le \rho_{bp} T_{bp}
 %   & \rho_{bp} = (1 + T_{fp} / T_{bp}) \rho - T_{fp} / T_{bp}
%\end{align}
%where $T_{bp}$ is the full backpropagation FLOPs. Although $T_{fp}/T_{bp}\approx 1/2$ generally holds for many NN models (e.g., MLPs), we still explicitly compute $T_{fp}/T_{bp}$ for a given LLM to ensure preciseness.

\noindent\textbf{Subproblem definition.}
As shown in Figure \ref{fig:subproblem_def}, we define each subproblem $P[k,t]$ as to maximize the cumulative importance of selected tensors when 1) selection is among the top $k$ tensors\footnote{We consider the tensor that is closest to the NN output as the topmost.} and 2) backpropagation FLOPs is at most $t$. DP starts by solving the smallest subproblem $P[k=1,t=1]$ and gradually solves larger subproblems based on the results of smaller subproblems and the recurrence relation of these subproblems, until the target problem $P[N, T_{full}]$ is solved. 

\noindent\textbf{Recurrence relations of subproblems.}
The recurrence relation between subproblem $P[k,t]$ and $P[k-1,t]$ depends on whether we further select the top tensor $k$ from the solution of $P[k-1,t]$, as shown in Figure \ref{fig:recurrence_relation}. \noindent\textbf{Case 1:} If $k$ is not selected, $P[k,t]$ will fall back to $P[k-1,t]$, since the importance of selected tensors will not be further increased.
\noindent\textbf{Case 2:} If $k$ is selected, then its FLOPs will be included into the solution of $P[k,t]$, no matter which other tensors are selected. The FLOPs involved with tensor $k$ include 1) the FLOPs to update tensor $k$ and 2) the FLOPs to pass activation gradients from the closest selected tensor $k_c$, such as tensor $k-3$ as shown in Figure \ref{fig:recurrence_relation}, to tensor $k$. This implies that $P[k,t]$ falls back to a previously solved subproblem $P[k-k_c,t-\Delta t]$, where
%\vspace{-0.05in}
\begin{equation} 
\Delta t = t_{dw_k}+\sum\nolimits_{j=k_c}^{k-1}t_{dy_j}.
\label{eq:recursion}
%\vspace{-0.05in}
\end{equation}
Since $k_c$ is unknown in advance, we backtrace the previously solved subproblems and explore all the possibilities of $k_c$ by reducing the depth of backpropagation from $k$, and the optimal solution to $P[k,t]$ is the one with the highest cumulative importance of selected tensors. Based on this recurrence relation, we can solve all subproblems by traversing the subproblem space. The time complexity of solving each subproblem is $O(N)$, and the overall time complexity of DP is $O(N^2T_{full})$.

%On the other hand, since the time complexity of DP increases quadratically with $N$, its computing cost could still be high when being applied to extremely big LLMs (e.g., GPT-3). In these cases, we could further leverage the existing parallelization techniques (e.g., multithreading \cite{tan2008improving}) and hardware accelerators (e.g., GPUs) to speed up DP.

\vspace{-0.1in}
\section{Experiments}
\vspace{-0.1in}
%Describe implementation and evaluation setups including models, datasets, and metrics. It would be better to emphasize here what question we are asking for the evaluation, e.g., does GreenTrainer performs better for larger models?

%We implemented GreenTrainer in PyTorch and conducted our experiments on a Lambda Cloud instance with a Nvidia H100 80GB GPU and 24 vCPUs. [this part will need to be revised if we later add experiment results using multiple GPUs.]

In our evaluation, we include decoder-only LLMs including OPT \citep{zhang2022opt} and BLOOMZ \citep{muennighoff2022crosslingual}, and an encoder-decoder LLM, namely FLAN-T5 \citep{chung2022scaling}, with LLM sizes ranging from 350M to 6.7B. Our experiments are mainly conducted using the following two datasets of abstractive summarization:
%All the model structures and pre-trained weights are downloaded from Hugging Face \citenum{wolf2019huggingface} model hub. 
\vspace{-0.1in}
\begin{itemize}
\item \textbf{SciTLDR} \citep{cachola2020tldr} is a dataset of 5.4K text summaries on 3.2K papers. It contains both author-written and expert-derived TLDRs, where the latter is collected by an annotation protocol that produces high-quality summaries with low annotation burden. %We let LLMs summarize the paper abstracts.
\vspace{-0.05in}
\item \textbf{DialogSum} \citep{chen2021dialogsum} is a dialogue summarization dataset of 13,460 dialogues with manually labeled summaries and topics. It has been demonstrated more challenging than other summarization datasets, such as SAMSum \citep{gliwa2019samsum} and CNN/Daily \citep{nallapati2016abstractive} at a similar scale.	
\end{itemize}
\vspace{-0.1in}

We also perform generative QA tasks on WebQuestion \citep{berant2013semantic} and PIQA \citep{bisk2020piqa} datasets in Appendix \ref{appendix:qa}. However, we do not consider non-generative tasks such as sentimental classification, entailment classification and extractive QA, because these tasks are too easy for LLMs and testing them with LLMs will result in exaggerated performance gain over the baseline.

For OPT and BLOOMZ, we follow GPT2-like prompt structures \citep{radford2019language}, ``[source seq.] TL;DR:'', for summarization tasks to preprocess input data. For FLAN-T5, we adopt the prompt structure ``summarize: [source seq.]'' used in the original T5 pre-training. We truncate the source sequences so that the length of every preprocessed input sequence is within 512 tokens. On the test data, we use a beam search size of 4, and set the maximum number of generated tokens to 64 for SciTLDR and 128 for DialogSum. We compare GreenTrainer (GT) with the following baselines:
\vspace{-0.25in}
\begin{itemize}
\item \textbf{Full Fine-Tuning (Full FT)} fine-tunes all the LLM parameters and should intuitively achieve the best accuracy of the trained model.
\vspace{-0.05in}
\item \textbf{Fine-Tuning Top2 (FT-Top2)} only fine-tunes the last two layers, typically the embedding layer and a LayerNorm. The input and output embedding layers are tied for OPT and BLOOMZ, but are not tied for FLAN-T5. This naive baseline only fine-tunes the smallest portion of LLM parameters and is used to identify whether the dataset is trivial to the LLM.  
\vspace{-0.2in}
\item \textbf{Prefix Tuning (Prefix-T)} \citep{li2021prefix} inserts trainable prefixes into each transformer block's input sequence while freezing the model parameters. For encoder-decoder LLMs, the trainable prefixes are only inserted into the decoder blocks.
\vspace{-0.05in}
\item \textbf{LoRA} \citep{hu2021lora} is currently the most popular method for efficient LLM fine-tuning. It uses low-rank matrix decomposition to reduce the training cost. We apply LoRA to both query and value projectors, as suggested in \citep{hu2021lora}.
\end{itemize}
\vspace{-0.1in}

In all experiments, we use a batch size of 4 and fine-tune the model for 5 epochs. We use the AdamW optimizer \citep{loshchilov2017decoupled} at a learning rate of $2\times10^{-5}$ with linear schedule and weight decay of $10^{-2}$. We use the ROUGE scores (\%R1/R2/RL) \citep{lin-2004-rouge} as the accuracy metric, and measure both Peta-FLOPs (PFLOPs) and wall-clock time as the training cost in each run. We measure the end-to-end cost of training, including the computing costs in forward and backward passes, and the computing costs of tensor importance evaluation and tensor selection using DP.

%We set the GreenTrainer's objective of FLOPs reduction to be (0.5, 0.7) for OPT and BLOOMZ models, and (0.34, 0.4, 0.5) for FLAN-T5, for pair-wise comparison with baselines at different levels of the training cost.
\vspace{-0.1in}
\subsection{Training Cost \& Accuracy}
\vspace{-0.1in}
We first evaluate the training cost and accuracy of GreenTrainer (GT). As shown in Table \ref{tab:cost_accuracy}, for the OPT-2.7B model, GT-0.5 can achieve the required 50\% of FLOPs reduction with at most 2\% accuracy loss, and GT-0.7 can even achieve 0.2\%-3\% higher ROUGE scores than Full FT. We hypothesize that this is because GT only fine-tunes the most important tensors and hence mitigates the possible overfitting in Full FT. Insufficient trainable parameters can also lead to underfitting, as FT-Top2 has significantly lower ROUGE scores. Similarly, compared to LoRA and Prefix Tuning, GT-0.7 achieves at least 2\% higher accuracy with the same amount of training FLOPs. 

%\vspace{-0.1in}
\begin{table}[ht]
	\centering
	{\fontsize{7}{8}\selectfont
		\begin{tabular}{lrrrrrr}
			\toprule
			\multirow{2}{*}{\makecell{\textbf{\# Model} \\ \textbf{\& Method}}} & \multicolumn{3}{c}{\textbf{SciTLDR}} & \multicolumn{3}{c}{\textbf{DialogSum}} \\
			\cmidrule(lr){2-4} \cmidrule(lr){5-7}
			& \textbf{PFLOPs} & \textbf{Time (h)} & \textbf{R1/R2/RL} & \textbf{PFLOPs} & \textbf{Time (h)} & \textbf{R1/R2/RL} \\
			\midrule
			\rowcolor{gray!25}
			\vspace{0.05in}
			\textbf{OPT-2.7B} \\
			Full FT        & 41.8  & 0.92  & 32.9/14.9/27.1   & 262.0  & 5.5  & 23.6/9.5/18.8   \\
			FT-Top2        & 29.0 (31\%$\downarrow$) & 0.61 (34\%$\downarrow$) & 9.1/4.0/7.6   & 181.6 (31\%$\downarrow$) & 3.8 (31\%$\downarrow$) & 20.8/7.9/17.5   \\
			Prefix-T        & 27.9 (33\%$\downarrow$) & 0.58 (37\%$\downarrow$) & 7.6/0.4/6.1   & 174.7 (33\%$\downarrow$)  & 3.7 (33\%$\downarrow$) & 13.4/3.3/10.9   \\
			LoRA        & 27.9 (33\%$\downarrow$)  & 0.59 (36\%$\downarrow$)  & 28.2/12.1/21.0   & 174.7 (33\%$\downarrow$) & 3.6 (35\%$\downarrow$) & 23.8/9.5/18.8   \\
			GT-0.5        & 20.8 (50\%$\downarrow$) & 0.46 (50\%$\downarrow$)  & 30.5/13.1/25.2  & 130.1 (50\%$\downarrow$) & 2.7 (51\%$\downarrow$) & 21.4/8.2/17.6   \\
			GT-0.7        & 29.2 (30\%$\downarrow$) & 0.68 (26\%$\downarrow$)  & 33.1/15.2/27.6   & 182.7 (30\%$\downarrow$) & 4.0 (27\%$\downarrow$) & 26.8/11.0/21.6   \\
			\midrule
			\rowcolor{gray!25}
			\vspace{0.05in}
			\textbf{BLOOMZ-3B} \\
			Full FT        & 47.2  & 1.0  & 28.3/12.1/22.5   & 294.8  & 6.5  & 26.1/10.6/21.0   \\
			FT-Top2        & 36.5 (23\%$\downarrow$) & 0.75 (25\%$\downarrow$) & 23.7/8.8/18.8   & 227.9 (23\%$\downarrow$) & 4.6 (29\%$\downarrow$) & 22.1/8.5/17.8    \\
			Prefix-T        & 31.5 (33\%$\downarrow$)  & 0.68 (34\%$\downarrow$) & 6.5/2.2/5.5   & 196.5 (33\%$\downarrow$) & 4.2 (35\%$\downarrow$) & 29.6/9.4/24.9   \\
			LoRA        & 31.5 (33\%$\downarrow$)  & 0.69 (33\%$\downarrow$) & 27.4/11.7/21.8   & 196.5 (33\%$\downarrow$) & 4.3 (34\%$\downarrow$)  & 35.4/14.3/28.6   \\
			GT-0.5        & 23.4 (51\%$\downarrow$)  & 0.51 (50\%$\downarrow$) & 26.7/10.7/21.2   & 146.4 (50\%$\downarrow$) & 3.1 (52\%$\downarrow$) & 24.9/9.5/20.0   \\
			GT-0.7        & 32.3 (32\%$\downarrow$) & 0.74 (28\%$\downarrow$) & 28.0/12.2/22.4   & 204.7 (31\%$\downarrow$) & 4.3 (34\%$\downarrow$) & 36.8/14.7/29.4   \\
			\midrule
			\rowcolor{gray!25}
			\vspace{0.05in}
			\textbf{FLAN-T5-3B} \\
			Full FT        & 21.7  & 0.64  & 37.1/18.5/31.7   & 135.7  & 4.0  & 46.5/20.8/38.5   \\
			FT-Top2        & 7.3 (66\%$\downarrow$) & 0.21 (67\%$\downarrow$)  & 36.5/18.4/31.5   & 46.1 (66\%$\downarrow$) & 1.4 (65\%$\downarrow$) & 39.2/16.7/32.9   \\
			Prefix-T        & 8.0 (63\%$\downarrow$) & 0.23 (64\%$\downarrow$) & 36.0/18.2/31.0   & 55.3 (60\%$\downarrow$)  & 1.7 (57\%$\downarrow$)  & 37.6/16.4/32.1   \\
			LoRA        & 14.4 (33\%$\downarrow$) & 0.41 (36\%$\downarrow$) & 36.6/18.5/31.5   & 90.5 (33\%$\downarrow$) & 2.5 (38\%$\downarrow$) & 44.7/19.8/37.1   \\
			GT-0.34        & 7.5 (65\%$\downarrow$) & 0.23 (64\%$\downarrow$) & 36.4/18.4/31.7   & 53.5 (61\%$\downarrow$) & 1.4 (65\%$\downarrow$) & 42.7/18.3/35.1   \\
			GT-0.4        & 10.0 (54\%$\downarrow$) & 0.38 (41\%$\downarrow$) & 36.7/18.5/31.5   & 62.5 (54\%$\downarrow$) & 2.3 (43\%$\downarrow$) & 46.0/20.7/38.1   \\
			GT-0.5       & 12.4 (43\%$\downarrow$) & 0.44 (31\%$\downarrow$) & 36.3/17.7/30.9   & 77.6 (43\%$\downarrow$) & 2.6 (35\%$\downarrow$) & 46.2/20.7/38.1   \\
			\bottomrule
	\end{tabular}}
\vspace{-0.05in}
	\caption{Comparison of the training cost \& accuracy in LLM fine-tuning. GreenTrainer with an objective $\rho$ of FLOPs reduction is denoted as GT-$\rho$.}
	\vspace{-0.15in}
	\label{tab:cost_accuracy}
\end{table}

Similarly, for BLOOMZ-3B, GT-0.5 can save 50\% training FLOPs and wall-clock time with $<2$\% accuracy loss. Compared to Full FT, GT-0.7 achieves the same ROUGE scores on SciTLDR, and 4\%-10\% higher on DialogSum. With the same amount of training FLOPs, GT-0.7 has 0.4\%-1.4\% higher ROUGE scores than LoRA. Note that both datasets are non-trivial for the BLOOMZ model, since the naive baseline (FT-Top2) still exhibits high accuracy loss.
%and Prefix-T performs much worse than other baselines on SciTLDR. The major reason may be that the inserted trainable prefixes break the original prompt structure and hence confuse the model.

%This is because FLAN-T5 has been instruction-fine-tuned after pre-training, and can potentially have better zero-shot adaptability.
For the FLAN-T5-3B model, FT-Top2 achieves similar fine-tuning qualities to Full FT with lower FLOPs, indicating that the SciTLDR dataset is trivial for FLAN-T5. In this case, GT-0.34 can achieve the same FLOPs and ROUGE scores by selecting a small portion of tensors. On the other hand, FT-Top2 loses accuracy significantly on DialogSum, but GT-0.4 reduces 54\% of training FLOPs and 43\% of wall-clock time without noticeable accuracy loss. GT-0.4 also outperforms LoRA by 1\% on ROUGE scores and reduces 11\% more FLOPs. Compared to Prefix tuning, GT-0.34 achieves 2\%-5\% higher ROUGE scores, while reducing the same amount of training FLOPs. 

\vspace{-0.1in}
\subsection{The Impact of FLOPs Reduction Objective}
\vspace{-0.1in}
To better understand how GreenTrainer performs with different objectives of FLOPs reduction, we vary the value of $\rho$ between 0.36 and 0.8, and compare GreenTrainer with LoRA on the OPT-2.7B model. As shown in Table \ref{tab:objective}, on the SciTLDR dataset, when the requirement of FLOPs reduction is high and corresponds to a value of $\rho\leq $0.4, GreenTrainer outperforms LoRA by achieving 2\% higher ROUGE scores and saving 25\% more FLOPs and wall-clock time. On the other hand, when the value of $\rho$ increases to 0.6, GreenTrainer outperforms the Full FT on ROUGE scores by 0.5\% and outperforms LoRA by 5.2\%, but saves 40\% of training FLOPs and 39\% of wall-clock time compared to Full FT. Similar results are also observed on the DialogSum dataset. In summary, with different objectives of FLOPs reduction, GreenTrainer can always provide better tradeoffs between the training accuracy and cost, compared to the SOTA baselines.

\begin{table}[ht]
	\vspace{-0.05in}
	\centering
	{\fontsize{7}{8}\selectfont
		\begin{tabular}{lrrrrrr}
			\toprule
			\multirow{2}{*}{\textbf{Method}} & \multicolumn{3}{c}{\textbf{SciTLDR}} & \multicolumn{3}{c}{\textbf{DialogSum}} \\
			\cmidrule(lr){2-4} \cmidrule(lr){5-7}
			& \textbf{PFLOPs} & \textbf{Time (h)} & \textbf{R1/R2/RL} & \textbf{PFLOPs} & \textbf{Time (h)} & \textbf{R1/R2/RL} \\
			Full FT        & 41.8  & 0.92  & 32.9/14.9/27.1   & 262.0  & 5.5  & 23.6/9.5/18.8   \\
			LoRA        & 27.9 (33\%$\downarrow$)  & 0.59 (36\%$\downarrow$)  & 28.2/12.1/21.0   & 174.7 (33\%$\downarrow$) & 3.6 (35\%$\downarrow$) & 23.8/9.5/18.8   \\
			GT-0.36        & 14.9 (64\%$\downarrow$)  & 0.32 (65\%$\downarrow$) & 4.1/1.7/3.6   & 92.9 (65\%$\downarrow$) & 1.9 (65\%$\downarrow$)  & 15.7/5.0/13.8   \\
			GT-0.4        & 16.6 (60\%$\downarrow$) & 0.36 (61\%$\downarrow$) & 28.6/11.6/23.5   & 103.4 (61\%$\downarrow$)  & 2.2 (60\%$\downarrow$) & 17.9/6.3/15.4   \\
			GT-0.5        & 20.8 (50\%$\downarrow$) & 0.46 (50\%$\downarrow$)  & 30.5/13.1/25.2  & 130.1 (50\%$\downarrow$) & 2.7 (51\%$\downarrow$) & 21.4/8.2/17.6   \\
			GT-0.6        & 25.0 (40\%$\downarrow$) & 0.56 (39\%$\downarrow$) & 33.4/15.3/27.8   & 156.6 (40\%$\downarrow$) & 3.3 (40\%$\downarrow$) & 24.0/9.7/19.2   \\
			GT-0.7        & 29.2 (30\%$\downarrow$) & 0.68 (26\%$\downarrow$)  & 33.1/15.2/27.6   & 182.7 (30\%$\downarrow$) & 4.0 (27\%$\downarrow$) & 26.8/11.0/21.6   \\
			GT-0.8        & 33.4 (20\%$\downarrow$) & 0.77 (16\%$\downarrow$) & 33.1/15.5/27.6   & 209.6 (20\%$\downarrow$) & 4.4 (20\%$\downarrow$) & 23.9/9.9/19.1   \\
			\bottomrule
	\end{tabular}}
	\vspace{-0.05in}
	\caption{Impact of different objectives of FLOPs reduction on the OPT-2.7B model}
%	\vspace{-0.1in}
	\label{tab:objective}
\end{table}

These results also demonstrate that GreenTrainer provides great flexibility in LLM fine-tuning between the training accuracy and cost, by adjusting the value of $\rho$. The user can opt to set a low value of $\rho$ ($\leq$0.4) to maximize the FLOPs reduction ($>$60\%) with moderate model accuracy loss (3\%-4\% on the two datasets we use). Alternatively, they can use a high value of $\rho$ ($\geq$0.6) to have the same level of FLOPs reduction as that of LoRA, but ensure the minimum model accuracy loss or even minor model accuracy improvement. We believe that such flexibility is practically important when fine-tuning LLMs for downstream tasks with different green AI requirements and constraints.

 \begin{table}[ht]
 	\vspace{-0.05in}
 	\centering
 	{\fontsize{8}{10}\selectfont
 		\begin{tabular}{lrrrrrr}
 			\toprule
 			\multirow{2}{*}{\textbf{Method}} & \multicolumn{3}{c}{\textbf{SciTLDR}} & \multicolumn{3}{c}{\textbf{DialogSum}} \\
 			\cmidrule(lr){2-4} \cmidrule(lr){5-7}
 			& \textbf{PFLOPs} & \textbf{Time (h)} & \textbf{R1/R2/RL} & \textbf{PFLOPs} & \textbf{Time (h)} & \textbf{R1/R2/RL} \\
 			Full FT        & 41.8  & 0.92  & 32.9/14.9/27.1   & 262.0  & 5.5  & 23.6/9.5/18.8   \\
 			GT-0.7 ($\Delta w$)        & 29.4 (30\%$\downarrow$) & 0.68 (26\%$\downarrow$)  & 32.7/15.2/27.2   & 183.8 (30\%$\downarrow$) & 4.0 (27\%$\downarrow$) & 24.9/10.2/19.7   \\
 			GT-0.7 ($\frac{\partial L}{\partial w}$)       & 29.4 (30\%$\downarrow$) & 0.67 (27\%$\downarrow$) & 32.8/15.1/27.2   & 184.0 (30\%$\downarrow$) & 4.0 (27\%$\downarrow$) & 25.0/10.2/20.0   \\
 			GT-0.7 ($\Delta w \frac{\partial L}{\partial w}$)  & 29.2 (30\%$\downarrow$) & 0.68 (26\%$\downarrow$)  & 33.1/15.2/27.6   & 182.7 (30\%$\downarrow$) & 4.0 (27\%$\downarrow$) & 26.8/11.0/21.6   \\
 			\bottomrule
 	\end{tabular}}
 	\vspace{-0.05in}
 	\caption{Efficacy of Tensor Importance Metrics (OPT-2.7B)}
 	\vspace{-0.15in}
 	\label{tab:importance_metric}
 \end{table}

 \vspace{-0.05in}
 \subsection{Efficacy of Tensor Importance Metrics}
 \vspace{-0.05in}
 The fine-tuning quality of GreenTrainer builds on proper evaluation of tensor importance. We compare our metric ($\Delta w \frac{\partial L}{\partial w}$) to the magnitude-based metric ($\Delta w$) \citep{lee2020layer} and the gradients-only metric ($\frac{\partial L}{\partial w}$) \citep{aji2017sparse}, using the OPT-2.7B model with $\rho=$0.7. As shown in Table \ref{tab:importance_metric}, with the same objective of FLOPs reduction, using our metric ($\Delta w \frac{\partial L}{\partial w}$) for tensor importance evaluation achieves the highest model accuracy and outperforms Full FT by 1\%-3\% on ROUGE scores. This is because magnitude-based metrics ignore the dependencies of weight updates. Gradient-only metrics only contain the direction information about tensor importance but cannot reflect the intensity of importance. Inaccurate importance measurements will in turn lead to inappropriate selections of trainable tensors.

\vspace{-0.1in}
\subsection{Impact of LLM Size}
\vspace{-0.1in}
A type of LLM may contain several variants with different sizes. To study GreenTrainer's performance with different LLM sizes, we performed fine-tuning using the OPT models with sizes ranging from 350M to 6.7B. As shown in Table \ref{tab:complexity}, even on small models (OPT-350M), GT-0.5 can save 17\%-21\% more training FLOPs than LoRA does, while achieving 2\%-4\% higher accuracy (on SciTDR) or the same accuracy (on DialogSum). When the model size increases to 2.7B, GT-0.5 outperforms LoRA and GT-0.7 outperforms Full FT on the SciTLDR dataset. On DialogSum, GT-0.7 performs similarly compared to LoRA. For the OPT-6.7B model\footnote{For the OPT-6.7B, Full FT and GT-0.7 with DialogSum have the out-of-memory issue on GPUs we use.}, GT-0.4 can save 27\% more training FLOPs than LoRA does on SciTLDR, while achieving the same model accuracy, and similar advantages can also be observed when comparing GT-0.5 and GT-0.7 with LoRA. Generally speaking, GreenTrainer's performance advantage widely applies to LLMs with different sizes.

\begin{table}[ht]
	\centering
		\vspace{-0.05in}
	{\fontsize{7}{8}\selectfont
		\begin{tabular}{lrrrrrr}
			\toprule
			\multirow{2}{*}{\makecell{\textbf{\# Params} \\ \textbf{\& Method}}} & \multicolumn{3}{c}{\textbf{SciTLDR}} & \multicolumn{3}{c}{\textbf{DialogSum}} \\
			\cmidrule(lr){2-4} \cmidrule(lr){5-7}
			& \textbf{PFLOPs} & \textbf{Time (h)} & \textbf{R1/R2/RL} & \textbf{PFLOPs} & \textbf{Time (h)} & \textbf{R1/R2/RL} \\
			\midrule
			\rowcolor{gray!25}
			\vspace{0.05in}
			\textbf{OPT-350M} \\
			Full FT        & 5.4  & 0.15  & 30.9/13.9/25.7   & 33.8  & 0.92  & 23.2/9.0/18.5   \\
			LoRA        & 3.6 (33\%$\downarrow$) & 0.10 (33\%$\downarrow$) & 25.9/10.8/20.3   & 22.5 (33\%$\downarrow$)  & 0.65 (29\%$\downarrow$)  & 21.5/7.7/17.3   \\ 
			GT-0.4      & 2.1 (61\%$\downarrow$) & 0.06 (60\%$\downarrow$) & 27.7/12.2/23.4   & 13.3 (61\%$\downarrow$) & 0.36 (61\%$\downarrow$) & 17.3/5.8/14.6   \\
			GT-0.5        & 2.7 (50\%$\downarrow$) & 0.08 (47\%$\downarrow$) & 29.9/13.2/24.9   & 16.7 (51\%$\downarrow$) & 0.45 (51\%$\downarrow$) & 21.3/7.8/17.3   \\
			GT-0.7        & 3.8 (30\%$\downarrow$) & 0.12 (20\%$\downarrow$) & 30.6/13.5/25.0   & 23.6 (30\%$\downarrow$) & 0.66 (28\%$\downarrow$) & 24.2/9.3/19.3   \\
			\midrule
			\rowcolor{gray!25}
			\vspace{0.05in}
			\textbf{OPT-1.3B} \\
			Full FT        & 20.8  & 0.46  & 32.1/14.3/26.4   & 130.8  & 2.9  & 25.4/10.3/20.2   \\
			LoRA        & 13.9 (33\%$\downarrow$) & 0.31 (33\%$\downarrow$)  & 28.1/11.9/22.0   & 87.2 (33\%$\downarrow$) & 1.9 (34\%$\downarrow$) & 24.6/9.9/19.4   \\
			GT-0.4      & 8.2 (61\%$\downarrow$) & 0.18 (61\%$\downarrow$) & 28.9/11.9/23.8   & 51.4 (61\%$\downarrow$) & 1.1 (62\%$\downarrow$) & 16.9/5.7/14.6   \\
			GT-0.5        & 10.3 (50\%$\downarrow$) & 0.23 (50\%$\downarrow$) & 30.0/12.7/24.5   & 64.2 (51\%$\downarrow$) & 1.4 (51\%$\downarrow$) & 20.1/7.4/16.7   \\
			GT-0.7        & 14.5 (30\%$\downarrow$) & 0.34 (26\%$\downarrow$) & 31.2/14.2/25.8   & 90.8 (30\%$\downarrow$) & 2.0 (31\%$\downarrow$) & 24.4/9.7/19.4   \\
			\midrule
			\rowcolor{gray!25}
			\vspace{0.05in}
			\textbf{OPT-2.7B} \\
			Full FT        & 41.8  & 0.92  & 32.9/14.9/27.1   & 262.0  & 5.5  & 23.6/9.5/18.8   \\
			LoRA        & 27.9 (33\%$\downarrow$)  & 0.59 (36\%$\downarrow$)  & 28.2/12.1/21.0   & 174.7 (33\%$\downarrow$) & 3.6 (35\%$\downarrow$) & 23.8/9.5/18.8   \\
			GT-0.4        & 16.6 (60\%$\downarrow$) & 0.36 (61\%$\downarrow$) & 28.6/11.6/23.5   & 103.4 (61\%$\downarrow$)  & 2.2 (60\%$\downarrow$) & 17.9/6.3/15.4   \\
			GT-0.5        & 20.8 (50\%$\downarrow$) & 0.46 (50\%$\downarrow$)  & 30.5/13.1/25.2  & 130.1 (50\%$\downarrow$) & 2.7 (51\%$\downarrow$) & 21.4/8.2/17.6   \\
			GT-0.7        & 29.2(30\%$\downarrow$) & 0.68 (26\%$\downarrow$)  & 33.1/15.2/27.6   & 182.7 (30\%$\downarrow$) & 4.0 (27\%$\downarrow$) & 26.8/11.0/21.6   \\
			\midrule
			\rowcolor{gray!25}
			\vspace{0.05in}
			\textbf{OPT-6.7B} \\
			Full FT        & 103.9  & 5.44  & 32.9/14.9/27.5   & 649.9  & -  & -   \\
			LoRA        & 69.3 (33\%$\downarrow$) & 1.3  & 28.4/12.3/22.7   & 433.3 (33\%$\downarrow$) & 8.1  & 24.9/10.2/19.4   \\
			GT-0.4        & 41.2 (60\%$\downarrow$) & 0.9  & 28.9/11.8/23.4   & 257.9 (60\%$\downarrow$)  & 5.2  & 19.7/7.0/16.3   \\
			GT-0.5        & 50.8 (51\%$\downarrow$) & 1.1  & 30.1/13.0/24.8   & 331.4 (49\%$\downarrow$)  & 6.7 & 21.8/8.5/17.3   \\
			GT-0.7        & 74.8 (28\%$\downarrow$) & 1.4  & 33.1/15.3/27.7  & -  & -  &  -  \\
			\bottomrule
	\end{tabular}}
	\vspace{-0.1in}
	\caption{Impact of LLM's model size}
	\vspace{-0.1in}
	\label{tab:complexity}
\end{table}

%\subsection{Behavior of Elastic Backpropagation}
\vspace{-0.1in}
\section{Conclusion}
\vspace{-0.15in}
In this paper, we present GreenTrainer, a new technique for LLM fine-tuning that allows efficient selection of trainable parameters via adaptive backpropagation, to ensure high training quality while minimizing the computation cost. GreenTrainer saves up to 64\% training FLOPs compared to full fine-tuning without noticeable accuracy loss. Compared to the existing technique such as Prefix Tuning and LoRA, GreenTrainer improves the accuracy by up to 4\% with the same FLOPs reduction.

%Although we use text summarization as the generative task in our evaluations, the rationale of GreenTrainer's adaptive backpropagation can also be applicable to large generative models in other applications, such as Stable Diffusion \citep{rombach2022high} for image generation and PaLM-E \citep{driess2023palm} for motion planning of multimodal embodied agents. Extensions to these domains will be our future work.

%Besides the table results below, we can also include (1) some "good" curve figures and (2) tensor selection heatmaps that are similar to Figure 21 in the ElasticTrainer paper.

\section*{Acknowledgments}
We would like to thank the anonymous reviewers and area chair for their comments and feedback. This work was supported in part by National Science Foundation (NSF) under grant number IIS-2205360, CCF-2217003 and CCF-2215042.

%\setcitestyle{numbers}
\bibliographystyle{abbrvnat}
\bibliography{ref}

\begin{thebibliography}{50}
\providecommand{\natexlab}[1]{#1}
\providecommand{\url}[1]{\texttt{#1}}
\expandafter\ifx\csname urlstyle\endcsname\relax
  \providecommand{\doi}[1]{doi: #1}\else
  \providecommand{\doi}{doi: \begingroup \urlstyle{rm}\Url}\fi

\bibitem[aii(2023)]{aiindex2023}
2023 {AI} index report.
\newblock \url{https://aiindex.stanford.edu/report/}, 2023.

\bibitem[Abadi(2016)]{abadi2016tensorflow}
M.~Abadi.
\newblock Tensorflow: learning functions at scale.
\newblock In \emph{Proceedings of the 21st ACM SIGPLAN International Conference
  on Functional Programming}, pages 1--1, 2016.

\bibitem[Aji and Heafield(2017)]{aji2017sparse}
A.~F. Aji and K.~Heafield.
\newblock Sparse communication for distributed gradient descent.
\newblock \emph{arXiv preprint arXiv:1704.05021}, 2017.

\bibitem[Ba et~al.(2016)Ba, Kiros, and Hinton]{ba2016layer}
J.~L. Ba, J.~R. Kiros, and G.~E. Hinton.
\newblock Layer normalization.
\newblock \emph{arXiv preprint arXiv:1607.06450}, 2016.

\bibitem[Berant et~al.(2013)Berant, Chou, Frostig, and
  Liang]{berant2013semantic}
J.~Berant, A.~Chou, R.~Frostig, and P.~Liang.
\newblock Semantic parsing on freebase from question-answer pairs.
\newblock In \emph{Proceedings of the 2013 conference on empirical methods in
  natural language processing}, pages 1533--1544, 2013.

\bibitem[Bisk et~al.(2020)Bisk, Zellers, Gao, Choi, et~al.]{bisk2020piqa}
Y.~Bisk, R.~Zellers, J.~Gao, Y.~Choi, et~al.
\newblock Piqa: Reasoning about physical commonsense in natural language.
\newblock In \emph{Proceedings of the AAAI conference on artificial
  intelligence}, volume~34, pages 7432--7439, 2020.

\bibitem[Breiman(2001)]{breiman2001random}
L.~Breiman.
\newblock Random forests.
\newblock \emph{Machine learning}, 45:\penalty0 5--32, 2001.

\bibitem[Brown et~al.(2020)Brown, Mann, Ryder, Subbiah, Kaplan, Dhariwal,
  Neelakantan, Shyam, Sastry, Askell, et~al.]{brown2020language}
T.~Brown, B.~Mann, N.~Ryder, M.~Subbiah, J.~D. Kaplan, P.~Dhariwal,
  A.~Neelakantan, P.~Shyam, G.~Sastry, A.~Askell, et~al.
\newblock Language models are few-shot learners.
\newblock \emph{Advances in neural information processing systems},
  33:\penalty0 1877--1901, 2020.

\bibitem[Cachola et~al.(2020)Cachola, Lo, Cohan, and Weld]{cachola2020tldr}
I.~Cachola, K.~Lo, A.~Cohan, and D.~S. Weld.
\newblock Tldr: Extreme summarization of scientific documents.
\newblock \emph{arXiv preprint arXiv:2004.15011}, 2020.

\bibitem[Candel et~al.(2023)Candel, McKinney, Singer, Pfeiffer, Jeblick,
  Prabhu, Gambera, Landry, Bansal, Chesler, et~al.]{candel2023h2ogpt}
A.~Candel, J.~McKinney, P.~Singer, P.~Pfeiffer, M.~Jeblick, P.~Prabhu,
  J.~Gambera, M.~Landry, S.~Bansal, R.~Chesler, et~al.
\newblock h2ogpt: Democratizing large language models.
\newblock \emph{arXiv preprint arXiv:2306.08161}, 2023.

\bibitem[Chen et~al.(2021)Chen, Liu, Chen, and Zhang]{chen2021dialogsum}
Y.~Chen, Y.~Liu, L.~Chen, and Y.~Zhang.
\newblock Dialogsum: A real-life scenario dialogue summarization dataset.
\newblock \emph{arXiv preprint arXiv:2105.06762}, 2021.

\bibitem[Chung et~al.(2022)Chung, Hou, Longpre, Zoph, Tay, Fedus, Li, Wang,
  Dehghani, Brahma, et~al.]{chung2022scaling}
H.~W. Chung, L.~Hou, S.~Longpre, B.~Zoph, Y.~Tay, W.~Fedus, E.~Li, X.~Wang,
  M.~Dehghani, S.~Brahma, et~al.
\newblock Scaling instruction-finetuned language models.
\newblock \emph{arXiv preprint arXiv:2210.11416}, 2022.

\bibitem[Devlin et~al.(2018)Devlin, Chang, Lee, and Toutanova]{devlin2018bert}
J.~Devlin, M.-W. Chang, K.~Lee, and K.~Toutanova.
\newblock Bert: Pre-training of deep bidirectional transformers for language
  understanding.
\newblock \emph{arXiv preprint arXiv:1810.04805}, 2018.

\bibitem[Gliwa et~al.(2019)Gliwa, Mochol, Biesek, and Wawer]{gliwa2019samsum}
B.~Gliwa, I.~Mochol, M.~Biesek, and A.~Wawer.
\newblock Samsum corpus: A human-annotated dialogue dataset for abstractive
  summarization.
\newblock \emph{arXiv preprint arXiv:1911.12237}, 2019.

\bibitem[Hesse et~al.(2021)Hesse, Schaub-Meyer, and Roth]{hesse2021fast}
R.~Hesse, S.~Schaub-Meyer, and S.~Roth.
\newblock Fast axiomatic attribution for neural networks.
\newblock \emph{Advances in Neural Information Processing Systems},
  34:\penalty0 19513--19524, 2021.

\bibitem[Hu et~al.(2021)Hu, Shen, Wallis, Allen-Zhu, Li, Wang, Wang, and
  Chen]{hu2021lora}
E.~J. Hu, Y.~Shen, P.~Wallis, Z.~Allen-Zhu, Y.~Li, S.~Wang, L.~Wang, and
  W.~Chen.
\newblock Lora: Low-rank adaptation of large language models.
\newblock \emph{arXiv preprint arXiv:2106.09685}, 2021.

\bibitem[Hu et~al.(2023)Hu, Lan, Wang, Xu, Lim, Lee, Bing, and
  Poria]{hu2023llm}
Z.~Hu, Y.~Lan, L.~Wang, W.~Xu, E.-P. Lim, R.~K.-W. Lee, L.~Bing, and S.~Poria.
\newblock Llm-adapters: An adapter family for parameter-efficient fine-tuning
  of large language models.
\newblock \emph{arXiv preprint arXiv:2304.01933}, 2023.

\bibitem[Huang et~al.(2023{\natexlab{a}})Huang, Yang, and
  Gao]{huang2023elastictrainer}
K.~Huang, B.~Yang, and W.~Gao.
\newblock Elastictrainer: Speeding up on-device training with runtime elastic
  tensor selection.
\newblock In \emph{Proceedings of the 21st Annual International Conference on
  Mobile Systems, Applications and Services}, pages 56--69, 2023{\natexlab{a}}.

\bibitem[Huang et~al.(2023{\natexlab{b}})Huang, Yang, and
  Gao]{huang2023modality}
K.~Huang, B.~Yang, and W.~Gao.
\newblock Modality plug-and-play: Elastic modality adaptation in multimodal
  llms for embodied ai.
\newblock \emph{arXiv preprint arXiv:2312.07886}, 2023{\natexlab{b}}.

\bibitem[Jin et~al.(2020)Jin, Yi, Zhang, Zhang, Schewe, and Huang]{jin2020does}
G.~Jin, X.~Yi, L.~Zhang, L.~Zhang, S.~Schewe, and X.~Huang.
\newblock How does weight correlation affect generalisation ability of deep
  neural networks?
\newblock \emph{Advances in Neural Information Processing Systems},
  33:\penalty0 21346--21356, 2020.

\bibitem[Kwon et~al.(2023)Kwon, Li, Venieris, Chauhan, Lane, and
  Mascolo]{kwon2023tinytrain}
Y.~D. Kwon, R.~Li, S.~I. Venieris, J.~Chauhan, N.~D. Lane, and C.~Mascolo.
\newblock Tinytrain: Deep neural network training at the extreme edge.
\newblock \emph{arXiv preprint arXiv:2307.09988}, 2023.

\bibitem[Lamb et~al.(2016)Lamb, ALIAS PARTH~GOYAL, Zhang, Zhang, Courville, and
  Bengio]{lamb2016professor}
A.~M. Lamb, A.~G. ALIAS PARTH~GOYAL, Y.~Zhang, S.~Zhang, A.~C. Courville, and
  Y.~Bengio.
\newblock Professor forcing: A new algorithm for training recurrent networks.
\newblock \emph{Advances in neural information processing systems}, 29, 2016.

\bibitem[Lee et~al.(2020)Lee, Park, Mo, Ahn, and Shin]{lee2020layer}
J.~Lee, S.~Park, S.~Mo, S.~Ahn, and J.~Shin.
\newblock Layer-adaptive sparsity for the magnitude-based pruning.
\newblock \emph{arXiv preprint arXiv:2010.07611}, 2020.

\bibitem[Lee et~al.(2018)Lee, Ajanthan, and Torr]{lee2018snip}
N.~Lee, T.~Ajanthan, and P.~H. Torr.
\newblock Snip: Single-shot network pruning based on connection sensitivity.
\newblock \emph{arXiv preprint arXiv:1810.02340}, 2018.

\bibitem[Lester et~al.(2021)Lester, Al-Rfou, and Constant]{lester2021power}
B.~Lester, R.~Al-Rfou, and N.~Constant.
\newblock The power of scale for parameter-efficient prompt tuning.
\newblock \emph{arXiv preprint arXiv:2104.08691}, 2021.

\bibitem[Li et~al.(2016)Li, Kadav, Durdanovic, Samet, and Graf]{li2016pruning}
H.~Li, A.~Kadav, I.~Durdanovic, H.~Samet, and H.~P. Graf.
\newblock Pruning filters for efficient convnets.
\newblock \emph{arXiv preprint arXiv:1608.08710}, 2016.

\bibitem[Li and Liang(2021)]{li2021prefix}
X.~L. Li and P.~Liang.
\newblock Prefix-tuning: Optimizing continuous prompts for generation.
\newblock \emph{arXiv preprint arXiv:2101.00190}, 2021.

\bibitem[Liao et~al.(2023)Liao, Tan, and Monz]{liao2023make}
B.~Liao, S.~Tan, and C.~Monz.
\newblock Make your pre-trained model reversible: From parameter to memory
  efficient fine-tuning.
\newblock \emph{arXiv preprint arXiv:2306.00477}, 2023.

\bibitem[Lin(2004)]{lin-2004-rouge}
C.-Y. Lin.
\newblock {ROUGE}: A package for automatic evaluation of summaries.
\newblock In \emph{Text Summarization Branches Out}, pages 74--81, Barcelona,
  Spain, July 2004. Association for Computational Linguistics.
\newblock URL \url{https://www.aclweb.org/anthology/W04-1013}.

\bibitem[Lin et~al.(2022)Lin, Zhu, Chen, Wang, Gan, and Han]{lin2022device}
J.~Lin, L.~Zhu, W.-M. Chen, W.-C. Wang, C.~Gan, and S.~Han.
\newblock On-device training under 256kb memory.
\newblock \emph{Advances in Neural Information Processing Systems},
  35:\penalty0 22941--22954, 2022.

\bibitem[Liu et~al.(2021)Liu, Zhang, Kuang, Zhou, Xue, Wang, Chen, Yang, Liao,
  and Zhang]{liu2021group}
L.~Liu, S.~Zhang, Z.~Kuang, A.~Zhou, J.-H. Xue, X.~Wang, Y.~Chen, W.~Yang,
  Q.~Liao, and W.~Zhang.
\newblock Group fisher pruning for practical network compression.
\newblock In \emph{International Conference on Machine Learning}, pages
  7021--7032. PMLR, 2021.

\bibitem[Liu et~al.(2022)Liu, Ji, Fu, Tam, Du, Yang, and Tang]{liu2022p}
X.~Liu, K.~Ji, Y.~Fu, W.~Tam, Z.~Du, Z.~Yang, and J.~Tang.
\newblock P-tuning: Prompt tuning can be comparable to fine-tuning across
  scales and tasks.
\newblock In \emph{Proceedings of the 60th Annual Meeting of the Association
  for Computational Linguistics (Volume 2: Short Papers)}, pages 61--68, 2022.

\bibitem[Loshchilov and Hutter(2017)]{loshchilov2017decoupled}
I.~Loshchilov and F.~Hutter.
\newblock Decoupled weight decay regularization.
\newblock \emph{arXiv preprint arXiv:1711.05101}, 2017.

\bibitem[Lu et~al.(2021)Lu, Grover, Abbeel, and Mordatch]{lu2021pretrained}
K.~Lu, A.~Grover, P.~Abbeel, and I.~Mordatch.
\newblock Pretrained transformers as universal computation engines.
\newblock \emph{arXiv preprint arXiv:2103.05247}, 1, 2021.

\bibitem[Malladi et~al.(2023)Malladi, Gao, Nichani, Damian, Lee, Chen, and
  Arora]{malladi2023fine}
S.~Malladi, T.~Gao, E.~Nichani, A.~Damian, J.~D. Lee, D.~Chen, and S.~Arora.
\newblock Fine-tuning language models with just forward passes.
\newblock \emph{arXiv preprint arXiv:2305.17333}, 2023.

\bibitem[Muennighoff et~al.(2022)Muennighoff, Wang, Sutawika, Roberts,
  Biderman, Scao, Bari, Shen, Yong, Schoelkopf,
  et~al.]{muennighoff2022crosslingual}
N.~Muennighoff, T.~Wang, L.~Sutawika, A.~Roberts, S.~Biderman, T.~L. Scao,
  M.~S. Bari, S.~Shen, Z.-X. Yong, H.~Schoelkopf, et~al.
\newblock Crosslingual generalization through multitask finetuning.
\newblock \emph{arXiv preprint arXiv:2211.01786}, 2022.

\bibitem[Nallapati et~al.(2016)Nallapati, Zhou, Gulcehre, Xiang,
  et~al.]{nallapati2016abstractive}
R.~Nallapati, B.~Zhou, C.~Gulcehre, B.~Xiang, et~al.
\newblock Abstractive text summarization using sequence-to-sequence rnns and
  beyond.
\newblock \emph{arXiv preprint arXiv:1602.06023}, 2016.

\bibitem[Ott et~al.(2019)Ott, Edunov, Baevski, Fan, Gross, Ng, Grangier, and
  Auli]{ott2019fairseq}
M.~Ott, S.~Edunov, A.~Baevski, A.~Fan, S.~Gross, N.~Ng, D.~Grangier, and
  M.~Auli.
\newblock fairseq: A fast, extensible toolkit for sequence modeling.
\newblock \emph{arXiv preprint arXiv:1904.01038}, 2019.

\bibitem[Paszke et~al.(2019)Paszke, Gross, Massa, Lerer, Bradbury, Chanan,
  Killeen, Lin, Gimelshein, Antiga, et~al.]{paszke2019pytorch}
A.~Paszke, S.~Gross, F.~Massa, A.~Lerer, J.~Bradbury, G.~Chanan, T.~Killeen,
  Z.~Lin, N.~Gimelshein, L.~Antiga, et~al.
\newblock Pytorch: An imperative style, high-performance deep learning library.
\newblock \emph{Advances in neural information processing systems}, 32, 2019.

\bibitem[Radford et~al.(2019)Radford, Wu, Child, Luan, Amodei, Sutskever,
  et~al.]{radford2019language}
A.~Radford, J.~Wu, R.~Child, D.~Luan, D.~Amodei, I.~Sutskever, et~al.
\newblock Language models are unsupervised multitask learners.
\newblock \emph{OpenAI blog}, 1\penalty0 (8):\penalty0 9, 2019.

\bibitem[Schwartz et~al.(2020)Schwartz, Dodge, Smith, and
  Etzioni]{schwartz2020green}
R.~Schwartz, J.~Dodge, N.~A. Smith, and O.~Etzioni.
\newblock Green ai.
\newblock \emph{Communications of the ACM}, 63\penalty0 (12):\penalty0 54--63,
  2020.

\bibitem[Scialom et~al.(2022)Scialom, Chakrabarty, and
  Muresan]{scialom2022fine}
T.~Scialom, T.~Chakrabarty, and S.~Muresan.
\newblock Fine-tuned language models are continual learners.
\newblock In \emph{Proceedings of the 2022 Conference on Empirical Methods in
  Natural Language Processing}, pages 6107--6122, 2022.

\bibitem[Sundararajan et~al.(2017)Sundararajan, Taly, and
  Yan]{sundararajan2017axiomatic}
M.~Sundararajan, A.~Taly, and Q.~Yan.
\newblock Axiomatic attribution for deep networks.
\newblock In \emph{International conference on machine learning}, pages
  3319--3328. PMLR, 2017.

\bibitem[Touvron et~al.(2023)Touvron, Lavril, Izacard, Martinet, Lachaux,
  Lacroix, Rozi{\`e}re, Goyal, Hambro, Azhar, et~al.]{touvron2023llama}
H.~Touvron, T.~Lavril, G.~Izacard, X.~Martinet, M.-A. Lachaux, T.~Lacroix,
  B.~Rozi{\`e}re, N.~Goyal, E.~Hambro, F.~Azhar, et~al.
\newblock Llama: Open and efficient foundation language models.
\newblock \emph{arXiv preprint arXiv:2302.13971}, 2023.

\bibitem[Vaswani et~al.(2017)Vaswani, Shazeer, Parmar, Uszkoreit, Jones, Gomez,
  Kaiser, and Polosukhin]{vaswani2017attention}
A.~Vaswani, N.~Shazeer, N.~Parmar, J.~Uszkoreit, L.~Jones, A.~N. Gomez,
  {\L}.~Kaiser, and I.~Polosukhin.
\newblock Attention is all you need.
\newblock \emph{Advances in neural information processing systems}, 30, 2017.

\bibitem[Wang and Gao(2023)]{wang2023tackling}
H.~Wang and W.~Gao.
\newblock Tackling the unlimited staleness in federated learning with
  intertwined data and device heterogeneities.
\newblock \emph{arXiv preprint arXiv:2309.13536}, 2023.

\bibitem[Wolf et~al.(2019)Wolf, Debut, Sanh, Chaumond, Delangue, Moi, Cistac,
  Rault, Louf, Funtowicz, et~al.]{wolf2019huggingface}
T.~Wolf, L.~Debut, V.~Sanh, J.~Chaumond, C.~Delangue, A.~Moi, P.~Cistac,
  T.~Rault, R.~Louf, M.~Funtowicz, et~al.
\newblock Huggingface's transformers: State-of-the-art natural language
  processing.
\newblock \emph{arXiv preprint arXiv:1910.03771}, 2019.

\bibitem[Zaken et~al.(2021)Zaken, Ravfogel, and Goldberg]{zaken2021bitfit}
E.~B. Zaken, S.~Ravfogel, and Y.~Goldberg.
\newblock Bitfit: Simple parameter-efficient fine-tuning for transformer-based
  masked language-models.
\newblock \emph{arXiv preprint arXiv:2106.10199}, 2021.

\bibitem[Zhang et~al.(2023)Zhang, Chen, Bukharin, He, Cheng, Chen, and
  Zhao]{zhang2023adaptive}
Q.~Zhang, M.~Chen, A.~Bukharin, P.~He, Y.~Cheng, W.~Chen, and T.~Zhao.
\newblock Adaptive budget allocation for parameter-efficient fine-tuning.
\newblock \emph{arXiv preprint arXiv:2303.10512}, 2023.

\bibitem[Zhang et~al.(2022)Zhang, Roller, Goyal, Artetxe, Chen, Chen, Dewan,
  Diab, Li, Lin, et~al.]{zhang2022opt}
S.~Zhang, S.~Roller, N.~Goyal, M.~Artetxe, M.~Chen, S.~Chen, C.~Dewan, M.~Diab,
  X.~Li, X.~V. Lin, et~al.
\newblock Opt: Open pre-trained transformer language models.
\newblock \emph{arXiv preprint arXiv:2205.01068}, 2022.

\end{thebibliography}

\newpage

\appendix

\section{Appendix}

\subsection{Reducing the memory usage of tensor importance evaluation}
\label{appendix:memory}
Our approach to evaluating the importance of NN tensors in Section 3.2 requires caching all the previous model weights and the current gradients, in order to compute Eq. (\ref{eq:importance}). However, doing so significantly increases the GPU memory consumption, especially for modern LLMs with billions of model weights. To reduce such GPU memory usage, we observe that our problem formulation in Eq. (\ref{eq:detailed_formulation}) will prevent tensors in early layers to be selected for training, due to the high costs of propagating their activation gradients in backpropagation. Hence, we could safely exclude these tensors from the trainable portion of LLM fine-tuning and save a significant amount of GPU memory. More specifically, the backpropagation during tensor importance evaluation can be early stopped at a certain tensor $k$, such that 
\begin{align}
\sum_{i=k-1,...,N} t_{dy_i} < \rho T_{full} \le \sum_{i=k,...,N} t_{dy_i},
\end{align}
i.e., the cumulative FLOPs of all the tensors from 1 to $k$ just exceeds our objective of FLOPs reduction. As shown in Table \ref{tab:reduce_memory}, by applying such early stopping method, we could proportionally save GPU memory with respect to the value of $\rho$, as a smaller value of $\rho$ leads to smaller $k$ and the backpropagation can hence be stopped earlier. For example, when $\rho=$50\%, 25\% of GPU memory can be saved, and such saving could further increase to 50\% when $\rho=$34\%.

\begin{table}[ht]
	\centering
	{\fontsize{8}{9}\selectfont
		\begin{tabular}{lccccc}
			\toprule
			\multirow{2}{*}{\textbf{Model}} 
			& \multirow{2}{*}{\makecell{\textbf{Full} \\ \textbf{evaluation}}}
			& \multirow{2}{*}{\makecell{\textbf{Early-stop} \\ \textbf{$\rho=34\%$}}}
			& \multirow{2}{*}{\makecell{\textbf{Early-stop} \\ \textbf{$\rho=40\%$}}}
			& \multirow{2}{*}{\makecell{\textbf{Early-stop} \\ \textbf{$\rho=50\%$}}}
			& \multirow{2}{*}{\makecell{\textbf{Early-stop} \\ \textbf{$\rho=60\%$}}} \\ \\
			%			& \multirow{2}{*}{\makecell{\textbf{CPU} \\ \textbf{offloading}}} 
			%			& \multirow{2}{*}{\makecell{\textbf{Max. time} \\ \textbf{overhead}}} \\ \\
			\midrule
			OPT-2.7B        & 10.8  & 5.5 & 6.5 & 8.1 & 9.7      \\
			FLAN-T5-3B        & 12.0  & 6.1 & 7.2 & 9.0 & 10.8     \\
			\bottomrule
	\end{tabular}}
	\caption{GPU memory consumption (in GigaBytes) of tensor importance evaluation}
	%\vspace{0.1in}
	\label{tab:reduce_memory}
\end{table}

\subsection{Reducing the computational cost of dynamic programming for tensor selection}
\label{appendix:dp}
In our proposed dynamic programming (DP) approach for tensor selection in Section 3.3, due to the high volume of FLOPs in LLM fine-tuning, the value of $T_{full}$ could be very large. To reduce the computational cost of DP, we can reduce the subproblem space by skipping two types of subproblems: 1) \textbf{invalid ones}, whose FLOPs constraint $t$ exceeds the desired constraint ($\rho T_{full}$); 2) \textbf{redundant ones}, whose FLOPs to pass activation gradients to the maximally allowed depth ($k$) exceeds $t$. Our preliminary experiment show that, doing so on an OPT model with $\rho_{bp}=50\%$ can reduce the number of subproblems by 5.5$\times$ without affecting the optimality of training.

\begin{table}[ht]
	\centering
	{\fontsize{8}{10}\selectfont
		\begin{tabular}{crrrrr}
			\toprule
			\textbf{Model} & \textbf{$T_q=1e1$}	& \textbf{$T_q=1e2$} & \textbf{$T_q=1e3$} & \textbf{$T_q=1e4$} & \textbf{$T_q=1e5$} \\
			\midrule
			OPT-2.7B        & 0.02/64.1/32.0  & 0.04/47.6/30.1  & 0.64/49.8/30.7  & 7.5/50.0/30.9  & 76.5/50.0/30.9    \\
			BLOOMZ-3B  &  0.0001/33.3/9.30 & 0.007/45.7/25.2 & 0.21/49.5/27.2 & 2.3/49.8/27.1 & 25.3/50.0/27.1  \\
			FLAN-T5-3B  & 0.04/64.9/36.5  & 0.25/57.1/36.5  & 3.5/55.3/36.7 & 41.8/51.8/36.7   & 449/50.0/36.7   \\
			\bottomrule
	\end{tabular}}
	\caption{The impact of DP resolution $T_q$ on fine-tuning OPT-2.7B, BLOOMZ-3B, and FLAN-T5-3B LLMs, on the SciTLDR dataset with $\rho=50\%$. Each triplet [a/b/c] presents a) the percentage of wall-clock time incurred by DP compared to full fine-tuning, b) the percentage of FLOPs after reduction compared to full fine-tuning, and c) the testing ROUGE-1 score, respectively.}
	%\vspace{0.1in}
	\label{tab:dp_resolution}
\end{table}

Besides, to further reduce the number of subproblems, we scale tensors' FLOPs $({t_{dw}},{t_{dy}})$ by multiplying a factor of $Z$:
\begin{equation} 
\widetilde{t_{dw}} = \left\lfloor t_{dw}\cdot{Z} \right\rfloor, \ \ \ \ \widetilde{t_{dy}} = \left\lfloor t_{dy}\cdot{Z} \right\rfloor, \ \ \ \  
\label{eq:downscaling}
\end{equation}
where $Z = \frac{T_q}{T_{full}}$ reduces the backropagation FLOPs to a resolution of $T_q<T_{full}$. The overall time complexity of DP is then reduced to $O(N^2T_q)$. On the other hand, such reduced resolution could increase the ambiguity in DP and affect the training quality. To investigate such tradeoff between the training quality and cost, we conducted preliminary experiments on multiple LLMs. Results in Table \ref{tab:dp_resolution} show that, for both OPT-2.7B and BLOOMZ-3B models, setting $T_q=1e3$ reduces the DP overhead to $<1$\% without affecting the training quality. Similarly, for FLAN-T5-3B, choosing $T_q=1e2$ can retain good training quality with negligible overhead. On the other hand, when $T_q$ is too small, the solution of DP could be inaccurate and hence result in ineffective reduction of the training FLOPs.

\begin{figure}[ht]
	\centering
	\vspace{-0.05in}
	\includegraphics[width=0.9\linewidth]{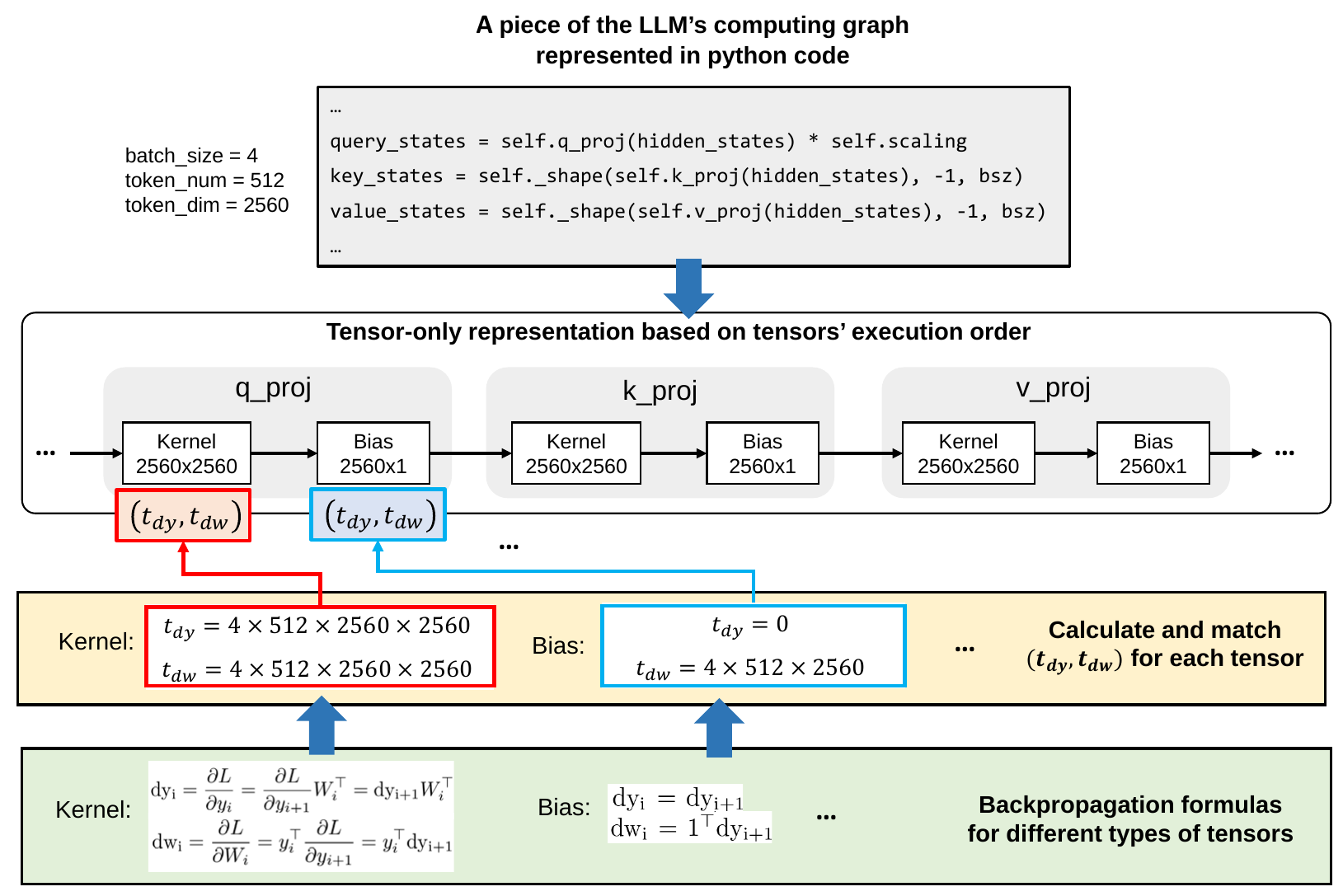}
	\vspace{-0.1in}
	\caption{An example of tensor FLOPs profiling in the OPT-2.7B model}
	\vspace{-0.05in}
	\label{fig:profiling_example}
	%	\vspace{-0.05in}
\end{figure}

\subsection{An example of tensor FLOPs profiling in the OPT-2.7B model}
\label{appendix:profiling}
To better facilitate understanding, we further show an example in Figure \ref{fig:profiling_example} about how we profile tensors in the OPT-2.7B models in our experiments. First, we convert the computing graph of the LLM, which is implemented in Python code, into a tensor-only representation. The tensors are ordered based on their execution orders in the forward pass, similar to the layer-level graph in Figure \ref{fig:flops_profiling}. We then calculate each tensor’s FLOPs ($t_{dy}$, $t_{dy}$) based on the backpropagation formulas discussed in Section 3.1. Such calculations are essentially counting the multiplications and being added in their formulas.

\begin{table}[ht]
	\centering
	\begin{minipage}{.5\linewidth}
		{\fontsize{8}{9}\selectfont
			\begin{tabular}{crrr}
				\toprule
				\textbf{Method} & \textbf{Accuracy (\%)}	& \textbf{PFLOPs} & \textbf{Time (h)} \\
				\midrule
				LoRA        & 49.5  & 174.0  & 6.27     \\
				GT-0.5        & 59.2  & 130.5  & 4.69     \\
				\bottomrule
		\end{tabular}}
		\caption{OPT-2.7B on PIQA dataset}
		\label{tab:piqa}
	\end{minipage}%
	\begin{minipage}{.5\linewidth}
		{\fontsize{8}{9}\selectfont
			\begin{tabular}{crrr}
				\toprule
				\textbf{Method} & \textbf{Accuracy (\%)}	& \textbf{PFLOPs} & \textbf{Time (h)} \\
				\midrule
				LoRA        & 19.6  & 16.0  & 0.55     \\
				GT-0.5        & 28.7  & 12.0  & 0.50     \\
				GT-0.6        & 29.5  & 14.0  & 0.61     \\
				\bottomrule
		\end{tabular}}
		\caption{OPT-2.7B on WebQuestion dataset}
		\label{tab:webqa}
	\end{minipage}%
	%\vspace{0.1in}
	\label{tab:piqa_webqa}
\end{table}

\subsection{Performance on generative question-answering tasks}
\label{appendix:qa}
To better evaluate the performance of GreenTrainer on other tasks, we also conducted experiments by using the OPT-2.7B model on WebQuestions and PIQA datasets for generative QA tasks. The WebQuestions dataset contains 6,642 QA pairs using Freebase as the knowledge base. The PIQA dataset focuses on multi-choice QA about physical knowledge with 21k QA pairs. We adopt the prompt format "\texttt{question:\{q\}</s>answer:\{a\}</s>}" for WebQuestions and "\texttt{goal:\{q\}</s>sol1:\{sol1\}</s>sol2:\{sol2\}</s>label:\{a\}</s>}" for PIQA, where \texttt{</s>} is the EOS token for OPT models. The hyper-parameters for training are the same as the ones described in Section 4. We evaluate the sentence-level accuracy which requires the generated answer to exactly match the ground truth. Note that for PIQA, the generated tokens are still predicted from the entire dictionary of OPT embeddings instead of from the two choices: the first or the second one. As shown in Table \ref{tab:piqa} and Table \ref{tab:webqa}, on both datasets, GreenTrainer (GT) achieves significantly better accuracy and time efficiency compared to LoRA.

In particular, the results on the PIQA dataset are generally lower than those reported in \cite{brown2020language}. The reason for this accuracy gap is that the way we use the OPT model to generate answers is more challenging than the setup in \cite{brown2020language}.  According to Section 2.4 in \cite{brown2020language}, it formulates the PIQA task as a multi-choice QA task where the answer is drawn from a small and predefined candidate set (e.g., [``0'', ``1'']), by comparing the probability scores only over the candidate tokens. In comparison, we strictly cast the problem to open-ended generation, where the candidate set is unknown. In that case, generating correct answers can be more difficult, because the model could generate totally irrelevant answers and increase its chance of making mistakes.

% \begin{table}[ht]
% 	\centering
% 	{\fontsize{8}{10}\selectfont
% 		\begin{tabular}{crrr}
% 			\toprule
% 			\textbf{Method} & \textbf{Accuracy (\%)}	& \textbf{PFLOPs} & \textbf{Time (h)} \\
% 			\midrule
%                 LoRA        & 19.6  & 16.0  & 0.55     \\
% 			GT-0.5        & 59.2  & 130.5  & 4.69     \\
% 			GT-0.6        & 49.5  & 174.0  & 6.27     \\
% 			\bottomrule
% 	\end{tabular}}
% 	\caption{OPT-2.7B on WebQuestion dataset}
% 	%\vspace{0.1in}
% 	\label{tab:webquestion}
% \end{table}

\subsection{Impact of frequency of tensor importance evaluation}
\label{appendix:importance}
Our design of GreenTrainer, by default, evaluates the importance of tensors and select the set of trainable tensors based on such importance at the beginning of each training epoch. Using the technical approach described in Section 3.1, such tensor importance evaluation is very lightweight, and our experiment results show that the overhead of importance evaluation is only 0.2\% on SciTLDR dataset and 0.01\% on DialogSum dataset, with respect to the entire fine-tuning cost.

On the other hand, in certain cases, the tensor importances, calculated from the model gradient changes, could exhibit non-negligible differences within one epoch. In these cases, the flexible design of GreenTrainer will allow us to adaptively increase the frequency of tensor importance evaluation and the corresponding DP-based tensor selection. To demonstrate the impact of such more frequent tensor importance evaluation and DP-based tensor selection, we conducted extra experiments using OPT-2.7B model on the WebQuestions dataset and generative QA task, as shown in Table \ref{tab:evaluation_freq}.

\begin{table}[ht]
	\centering
	{\fontsize{8}{10}\selectfont
		\begin{tabular}{crr}
			\toprule
			\textbf{Frequency of tensor importance evaluation} & \textbf{Accuracy (\%)}	& \textbf{Time (h)} \\
			\midrule
                Every 945 iterations (once per epoch)      & 28.4  & 0.50      \\
			Every 600 iterations       & 28.5  & 0.54     \\
			Every 400 iterations       & 28.2  & 0.56     \\
                Every 200 iterations       & 27.5  & 0.64     \\
			\bottomrule
	\end{tabular}}
	\caption{Impact of tensor importance evaluation frequency}
%	\vspace{-0.1in}
	\label{tab:evaluation_freq}
\end{table}

The results show that: (1) More frequent tensor importance evaluation brings only very small improvement on task accuracy. Considering the randomness in different training trials, we believe that such accuracy improvement is negligible, and the accuracy could even drop down by 1\% when the frequency of evaluation is very high (every 200 iterations). We believe that this is due to accumulation of tensor importance evaluation and tensor selection errors, which stem from the first-order approximation in the tensor importance metric and the approximate solution in DP. Another possible reason is that the tensor importances are calculated over the training dataset, and too frequent tensor importance evaluation may make the training process overfit to the training dataset. (2) The training cost steadily increases with the frequency of tensor importance evaluation. When the interval of evaluation reduces from 945 iterations to 200 iterations, the training time increases by 28\%.

In summary, performing more frequent tensor importance evaluation within each epoch brings little improvement on the task accuracy but noticeably increase the training cost. We believe that the tensor importances being evaluated once in each epoch would be sufficiently accurate for appropriate selection of trainable tensors.

\subsection{The necessity of dynamic tensor selection}
\label{appendix:dynamic}
If the LLM fine-tuning uses a fixed training dataset, it is possible that using a fixed tensor selection decided at the initial phase of training may not result in a significant model accuracy drop, compared to runtime tensor selection. However, in practical LLM fine-tuning scenarios, this assumption usually does not hold due to the following two reasons. First, in a lot of LLM fine-tuning scenarios, such as online learning and model personalization, the model is continuously retrained using online data, which is continuously generated at runtime with variant data distributions. Such variant data distributions will surely result in different importances of tensors through the training procedure and hence require runtime tensor selection. Such online LLM fine-tuning scenarios recently become more and more popular, especially with the possibility of deploying LLMs onto user's personal mobile devices such as smartphones. Second, even for a fixed training dataset, it is also possible that the importances of some tensors may change as the training progresses. In these cases, dynamic tensor selection could improve the trained model accuracy. To verify this, we conducted additional experiments using the OPT-2.7B model on the WebQuestions dataset and generative QA task. As shown in Table \ref{tab:selection_strategy}, dynamic tensor selection could make non-negligible contributions to improving the task accuracy, with negligible increase of training cost.

\begin{table}[ht]
	\centering
	{\fontsize{8}{10}\selectfont
		\begin{tabular}{crr}
			\toprule
			\textbf{Strategy} & \textbf{Accuracy (\%)}	& \textbf{Time (h)} \\
			\midrule
                Fixed tensor selection only in the first epoch of training      & 27.4  & 0.49      \\
			Dynamic tensor selection, once in each epoch       & 28.4  & 0.50     \\
			More frequent tensor selection (5 times in each epoch)       & 27.5  & 0.64     \\
			\bottomrule
	\end{tabular}}
	\caption{Different strategies of tensor selection}
%	\vspace{-0.1in}
	\label{tab:selection_strategy}
\end{table}

Note that, such improvement of model accuracy would be dependent on the specific dataset and model being used, but these experiment results above demonstrated the necessity of runtime tensor selection. In addition, our experiment results also showed that such tensor importance evaluation and selection indeed incur very little extra computing overhead.

\end{document}